%% file: main.tex
\documentclass[9.5pt,journal,compsoc]{IEEEtran}
\usepackage[T1]{fontenc}
\usepackage[utf8]{inputenc}

\ifCLASSOPTIONcompsoc
\usepackage[nocompress]{cite}
\else
\usepackage{cite}
\fi
\ifCLASSINFOpdf
\usepackage[pdftex]{graphicx}
\else
\usepackage[dvips]{graphicx}
\DeclareGraphicsExtensions{.eps}
\fi
\usepackage{tikz}
\usepackage{geometry}
\usepackage{amsmath,amsfonts,amssymb}
\usepackage{graphicx}
\usepackage{titlesec}
\usepackage{enumitem}
\usepackage{xcolor}
\usepackage{forest}
\usepackage{booktabs}
\usepackage{multirow}
\usepackage{hyperref}
\usepackage{longtable}
\usepackage{xltabular}
\usetikzlibrary{
    matrix,
    positioning, 
    arrows.meta, 
    backgrounds, 
    fit, 
    calc
}

\usepackage{amssymb}
\usepackage{amsthm}
\usepackage{array}
\usepackage{url}
\usepackage[linesnumbered,ruled]{algorithm2e}

\DeclareMathOperator*{\argmax}{argmax}

\theoremstyle{definition}

\let\oldnl\nl
\newcommand{\nonl}{\renewcommand{\nl}{\let\nl\oldnl}}
\renewcommand\thesection{\Alph{section}}
\usepackage{multirow}
\usepackage[font=scriptsize]{subfig}
\usepackage{epstopdf}
\usepackage{subfiles}
\usepackage{makecell}
\usepackage{ragged2e}
\usepackage{soul}
\soulregister{\ref}7
\soulregister{\cite}7
\usepackage{color, xcolor}

\usepackage{pifont}
\usepackage{xspace} 

\usepackage{verbatim}

\soulregister\cite7 
\soulregister\eqref7
\usepackage{colortbl}

\usepackage{algorithmic}

\usepackage{textcomp}
\usepackage{stfloats}
\usepackage{tabularx}
\usepackage{stmaryrd}
\usepackage{enumitem}
\usepackage{wrapfig}
\usepackage{etoolbox}
\usepackage{diagbox}
\usepackage{lettrine}
\usepackage{amssymb}
\usepackage{longtable}

\definecolor{improvegreen}{RGB}{0, 128, 0} 

\definecolor{improvered}{RGB}{220, 50, 50}
\setlist[itemize]{leftmargin=10pt}

\newcommand{\webref}[4][official page]{#2\footnote{\href{#3}{#1}. Accessed #4.}}

\newcommand{\showdelete}{1}
\DeclareRobustCommand{\del}[1]{%
  \if\showdelete0%
    {\begingroup\color{improvered}#1\endgroup}%
  \fi%
}
\DeclareRobustCommand{\add}[1]{%
  \if\showdelete0%
    {\begingroup\color{blue}#1\endgroup}%
  \else%
    {#1}%
  \fi%
}

\begin{document}


\title{Aligning Perception, Reasoning, Modeling and Interaction: A Survey on Physical AI}
\author{
Kun Xiang\IEEEauthorrefmark{1}, 
Terry Jingchen Zhang\IEEEauthorrefmark{1}, 
Yinya Huang\IEEEauthorrefmark{1}, 
Jixi He, 
Zirong Liu, 
Yueling Tang,\\ 
Ruizhe Zhou, 
Lijing Luo, 
Youpeng Wen, 
Xiuwei Chen, 
Bingqian Lin,
Jianhua Han,\\
Hang Xu,
Hanhui Li,
Bin Dong, 
Xiaodan Liang\IEEEauthorrefmark{2},~\IEEEmembership{Senior Member,~IEEE}

\IEEEcompsocitemizethanks{
	\IEEEcompsocthanksitem 
	\IEEEauthorrefmark{1}These three authors contribute equally to this work.
        \vspace{0.2cm}
	\IEEEcompsocthanksitem 
	\IEEEauthorrefmark{2}Xiaodan Liang is the corresponding author.
        \vspace{0.2cm}
        \IEEEcompsocthanksitem Kun Xiang, Jixi He, Zirong Liu, Yueling Tang, Ruizhe Zhou, Lijing Luo, Youpeng Wen, Xiuwei Chen, and Hanhui Li are with the Shenzhen Campus of Sun Yat-sen University, Shenzhen, China. 
        \protect\\ E-mail: \{xiangk@mail2.sysu.edu.cn\}
        \IEEEcompsocthanksitem Terry Jingchen Zhang is with ETH Zurich, Zurich, Switzerland. 
        \IEEEcompsocthanksitem Yinya Huang is a postdoctoral fellow at the ETH AI Center, ETH Zurich, Zurich, Switzerland. 
        \IEEEcompsocthanksitem Youpeng Wen is with The Chinese University of Hong Kong, Hong Kong.
        \IEEEcompsocthanksitem Bingqian Lin is a postdoc researcher with Shanghai Jiao Tong University, Shanghai, China.
        \IEEEcompsocthanksitem Hang Xu and Jianhua Han are with Yinwang Intelligent Technology Co., Ltd., Shenzhen, China. 
        \IEEEcompsocthanksitem Bin Dong is with Peking University and Beijing International Center for Mathematical Research, Beijing, China. 
        \IEEEcompsocthanksitem Xiaodan Liang is with the Shenzhen Campus of Sun Yat-sen University, Shenzhen, China. 
        \protect\\E-mail: \{liangxd9@mail.sysu.edu.cn\}
	}
}

\IEEEtitleabstractindextext{
\begin{abstract} \justifying
\input{sec/0_abstract}
\end{abstract}
\begin{IEEEkeywords} \justifying
Physical AI System, Physical Perception, Physics Reasoning, World Modeling, Embodied Interaction.
\end{IEEEkeywords}
}
\maketitle

\begin{quote}

\textit{``What I cannot create, I do not understand.''}

\hfill---\textsc{Richard Feynman}\footnote{Richard Feynman (1918-1988) was a Nobel Prize-winning physicist known for his groundbreaking work in quantum electrodynamics and the invention of Feynman diagrams.}
\end{quote}
\input{sec/1_intro}

\input{sec/2_preliminaries}
\input{sec/3_physical_reasoning}

\input{sec/4_physics_reasoning}
\input{sec/5_world_models}

\input{sec/6_embody_ai}
\input{sec/7_discussion}

\input{sec/8_conclusion}

    \tiny
    \bibliographystyle{IEEEtran}
    \bibliography{main}
\clearpage
\normalsize
\input{sec/x_appendix}

\end{document}

%% file: sec/0_abstract.tex
The convergence of embodied intelligence and world models has catalyzed growing interest in integrating physical laws into AI systems. While prior surveys have examined world models and embodied intelligence separately, we focus on the progression that connects these capabilities as a unified developmental pathway from passive observation to active physical comprehension. This survey provides a systematic framework revealing how physical AI advances through four interconnected stages: perception transforms sensory data into structured physical representations, reasoning derives explanations from observed phenomena, modeling enables predictive simulation grounded in physical principles, and embodied interaction closes the loop through physical manipulation and environmental feedback. Each stage enables and enhances the next: perceptual grounding supports causal reasoning, reasoning unlocks predictive capabilities, and robust models drive genuine physical interaction. Through analysis of developments spanning architectural innovations, training methodologies, causal inference, and embodied systems, we synthesize how physical understanding emerges through cumulative integration across this progression. Our framework reveals the evolution from isolated, task-specific solutions toward integrated architectures that advance from pattern recognition toward causal reasoning and counterfactual prediction. This perspective provides foundations for next-generation physical AI systems with direct implications for safe, generalizable, and interpretable deployment across robotics, scientific discovery, and autonomous systems. We maintain a continuously updated taxonomy repository at \href{https://github.com/AI4Phys/Awesome-AI-for-Physics}{https://github.com/AI4Phys/Awesome-AI-for-Physics}.

%% file: sec/1_intro.tex
\section{Introduction}

\lettrine[lines=2, loversize=0.1]{T}{eaching} artificial intelligence to understand our physical world represents one of the most fundamental challenges in modern AI research~\cite{World_models2018,AIMeetPhy_survey_2024}. While humans naturally grasp complex physical interactions from early childhood, frontier models struggle with basic physical reasoning that young children master effortlessly~\cite{NewtonianIU,Battaglia2016InteractionNF}. This capability gap becomes increasingly critical as AI systems are deployed in real-world scenarios ranging from self-driving vehicles to humanoid robots. To quantify this limitation, evaluation frameworks such as SeePhys~\cite{seephys} for symbolic reasoning, PHYRE~\cite{phyre,I-PHYRE} for intuitive physical perception, and PhyBlock~\cite{Ma2025PhyBlockAP} for realistic dynamics prediction have emerged, consistently demonstrating that current models lack structured understanding of physical laws. This gap manifests dramatically in performance disparities. A vision model trained on millions of images may achieve superhuman image classification accuracy yet fail to predict elementary outcomes such as a bouncing ball's trajectory~\cite{LLMPhy_2025,ComPhy}. This paradox reveals that contemporary AI models learn statistical correlations from massive data rather than developing causal knowledge rooted in physical principles.

\begin{figure*}
\centering
\includegraphics[width=\textwidth]{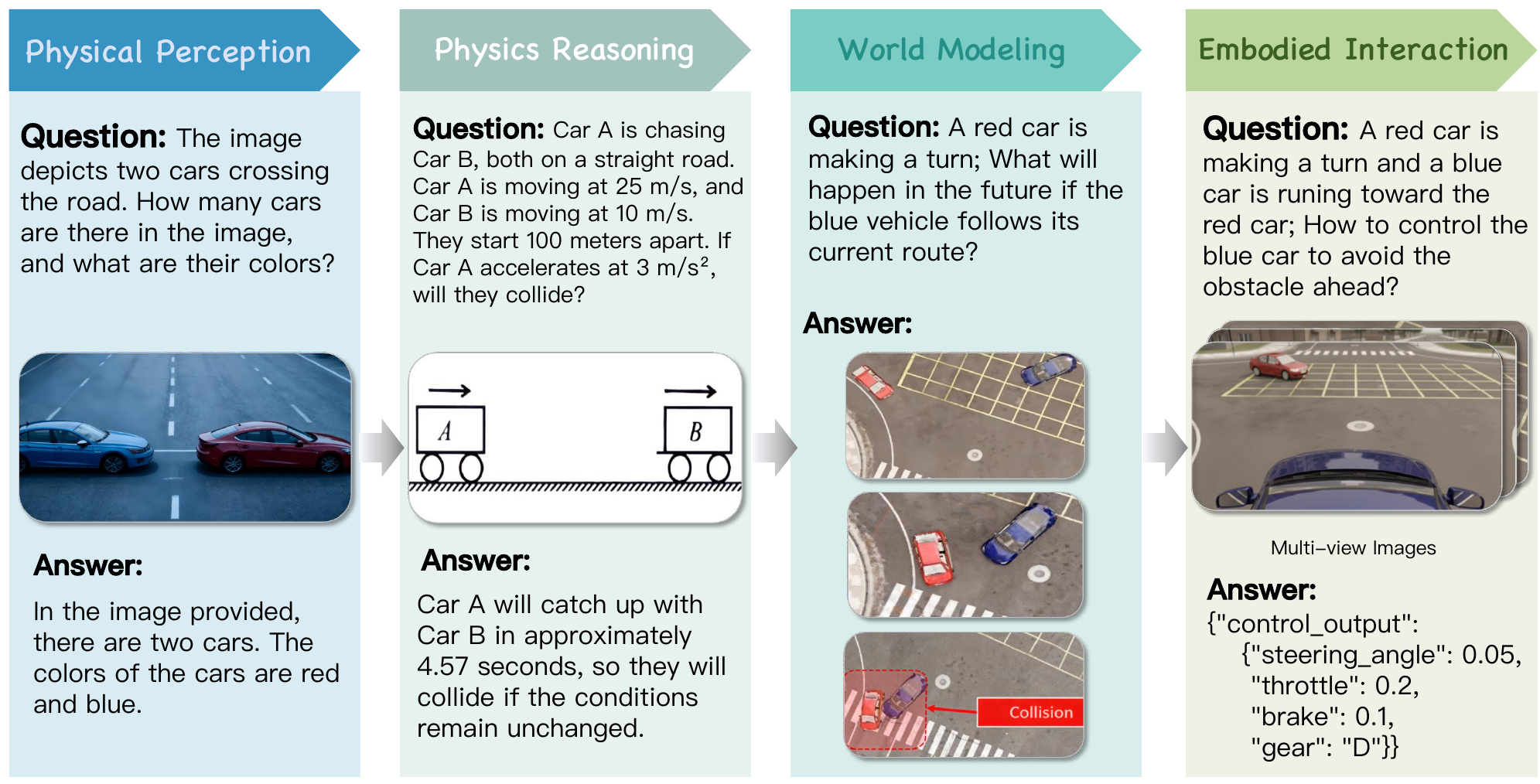}
\caption{Overview of four physical understanding capabilities of current AI systems. \textbf{Physical Perception} denotes the model's ability to recognize the properties of objects. \textbf{Physics Reasoning} refers to the model's capability to understand symbolic systems and physical laws. \textbf{World Modeling} represents its capacity for simulation and prediction. \textbf{Embodied Interaction} characterizes the model's ability to interact with the physical environment.}
\label{fig:overview}
\end{figure*}

Addressing this fundamental challenge requires physics-aware reasoning models across multiple complementary directions. From an \textit{architectural perspective}, Graph Neural Networks (GNNs)~\cite{gcn,gat,graphsage} leverage their inherent capacity to model relational structures and capture pairwise physical interactions between objects~\cite{Battaglia2016InteractionNF,VIN,NPE}, naturally encoding spatial relationships and interaction dynamics. Diffusion-based models~\cite{MotionCraft,VideoREPA} have demonstrated substantial potential in modeling complex physical processes through learned denoising procedures, while transformer architectures~\cite{Han2025APMA}\del{\cite{Kantamneni2024HowDTA}} increasingly incorporate physics-aware attention mechanisms and constraints. Beyond architecture, \textit{training methodology} plays a crucial role. Researchers have developed physics-guided loss functions that explicitly incorporate physical constraints~\cite{elhasadi2019_semisupervised_sr_drag,sun2020_surrogate_fluid_flow}, curriculum learning strategies that progressively introduce physical complexity~\cite{chen2025mint}, and reinforcement learning frameworks with physics-informed reward structures~\cite{AlphaDrive, AutoDrive,liu2025reinforcementlearningmeetslarge}. These training paradigms aim to instill physical intuition during learning rather than relying solely on pattern recognition. At \textit{test-time inference}, approaches explicitly integrate physical laws and symbolic reasoning into causal modeling~\cite{causalmodeling,causalthreads,PhysORD2024}, while differentiable physics engines enable end-to-end optimization~\cite{differentiable-stokes-flow2020, Scalable-differentiable-physic2020}, thereby bridging the gap between learned representations and established physical principles.

\begin{figure}[t]
    \centering
    \includegraphics[width=\columnwidth]{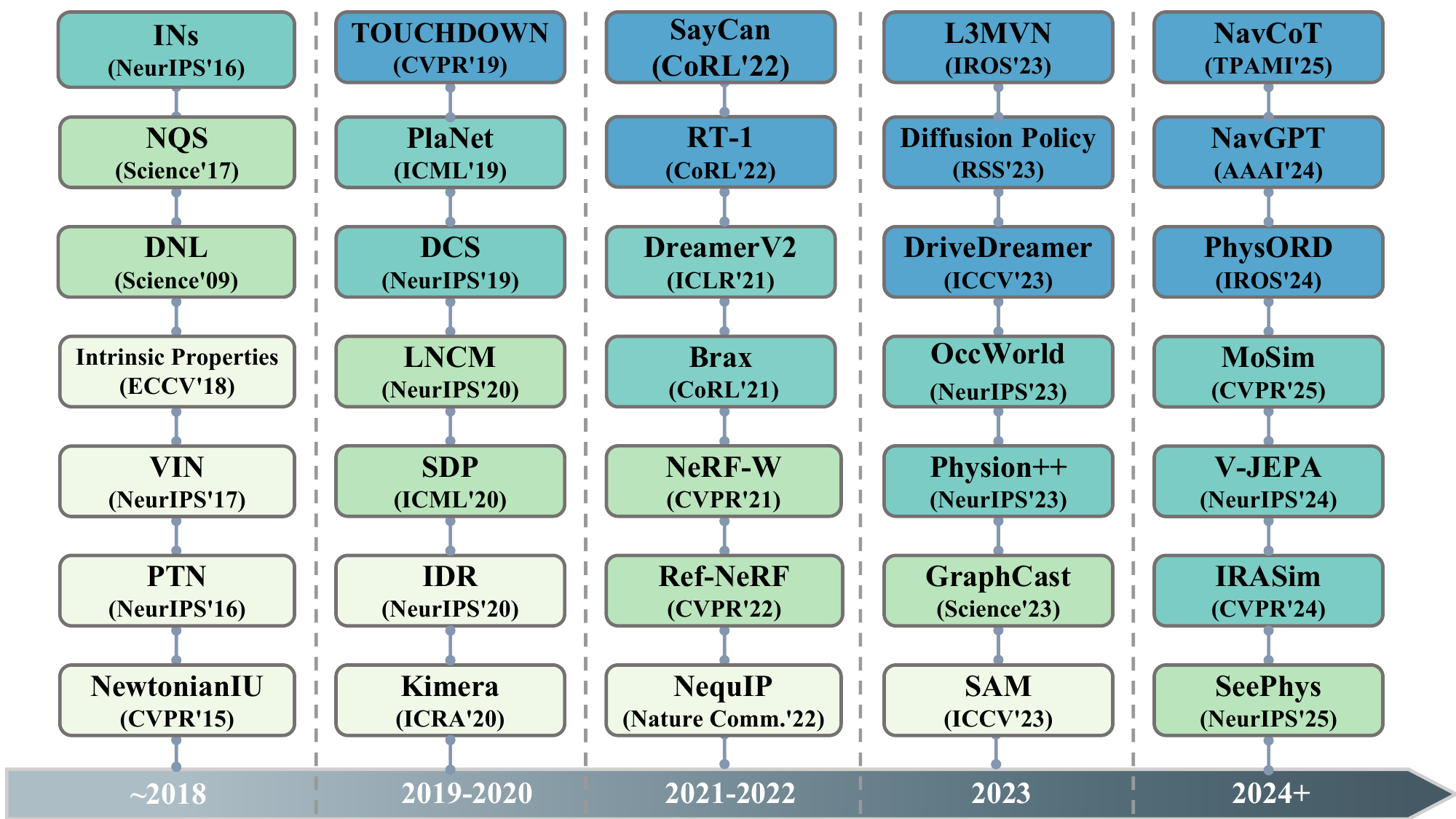}
    \caption{Timeline of the development of Physical AI Systems. The timeline illustrates the evolution of key research works, organized into chronological periods. The colored blocks indicate the primary research theme of each paper: pale mint for \textit{Physical Perception}, soft green for \textit{Physics Reasoning}, cyan for \textit{World Modeling}, and sky blue for \textit{Embodied Interaction}.}
    \label{fig:roadmap}
\end{figure}

Among these directions, large language models (LLMs) and multimodal large language models (MLLMs) offer a particularly promising pathway forward. These models have demonstrated substantial capability in learning from massive datasets and performing sophisticated end-to-end reasoning across multiple modalities~\cite{gpt4o,llava,o1,Qwen2.5-VL}, including \webref[Claude 3.7 Sonnet system card]{Claude 3.7 Sonnet}{https://www-cdn.anthropic.com/9ff93dfa8f445c932415d335c88852ef47f1201e.pdf}{Apr. 13, 2026} and \webref[Gemini 2.5 Pro model card]{Gemini 2.5 Pro}{https://storage.googleapis.com/deepmind-media/Model-Cards/Gemini-2-5-Pro-Model-Card.pdf}{Apr. 13, 2026}. \del{Multi-agent systems with tool integration have achieved performance comparable to top human competitors in physics competitions~\cite{Liang2025MultimodalRF},}\add{Tool-augmented agentic systems have achieved performance comparable to top human competitors in physics competitions~\cite{PhysicsSupernova}, while challenge-oriented multimodal reasoning systems have shown strong results on physics reasoning benchmarks~\cite{Liang2025MultimodalRF},} suggesting potential for more general physical understanding. Generative systems~\cite{GAIA-1,DriveDreamer} have achieved controllable, photorealistic synthesis of physical scenarios, while vision-language-action models~\cite{KimEtAl2024_OpenVLA,BlackEtAl2024_pi0} integrate natural language instructions with continuous physical manipulation, bridging the gap from abstract reasoning to embodied interaction. These developments reflect a broader paradigm shift in physical AI from isolated, task-specific solutions toward integrated architectures (Figure~\ref{fig:roadmap}) with synergies across domains, extending beyond recognizing input features toward generating and interacting with realistic physical scenarios.

\noindent\textbf{Scope Comparison and Contributions.}
As summarized in Table~\ref{tab:comparison}, existing surveys have examined individual dimensions of physical understanding in isolation, addressing perception~\cite{Yin2023ASO,Sapkota2025ObjectDW,Guo2024PhyGraspGR,Sun2025ProbingPC}\del{\cite{Marjieh2023LargeLM}}, reasoning~\cite{Zhou2025FromPT,Li2025FromS1,Ravishankara2025TheAI,Duan2022ASO,Huang2022TowardsRI}, modeling~\cite{Khan2025FoundationMD, Barman2025LargePM, Ding2024UnderstandingWO, Liu2025GenerativePA, Zhu2024IsSA, Kong20253DA4, Bergen2019MachineLF, Xie2025From2T, Chen2025ASO,Mai2024FromEM}, and interaction~\cite{Liu2024AligningCS, Qi2024ShapeLLMU3, Sathyam2025FoundationMF, Feng2025EmbodiedAF, Long2025ASL, Wang2023ASO, Liang2025LargeME, Zheng2024ASO, Wang2025TowardEA,Duan2021ASOA,Xu2024ASOA,Han2023ASOA} as separate research areas without examining the synergistic connections between them. Our survey uniquely focuses on the evolutionary trajectory that unites these four capabilities into a coherent paradigm, analyzing how they interact and inform one another toward unified physical AI systems. We examine how deep learning systems leverage the laws of physics to solve physics problems in an end-to-end manner, as opposed to how physics principles inspire neural network architectures. We leave physics-inspired architectures such as Boltzmann Machines and Hopfield Networks, as well as the broader field of deep learning for physics research, for dedicated AI4Science surveys. To this end, we present a three-tier taxonomy that systematically organizes research across these four fundamental capabilities (Figure~\ref{fig:overview}), with hierarchical task structures and detailed methodological analysis for each. Drawing from over 300 papers, our coverage spans advanced physics problems to applied tasks including object recognition, spatial perception, video generation, robotic control, and autonomous driving, revealing the integrated development pathway from passive observation to active physical comprehension. \add{For transparency and retrieval convenience, the appendix further provides a PRISMA-style survey construction record together with taxonomy lookup tables for methodological and benchmark papers.}

\noindent\textbf{Survey Structure.} 
Section~\ref{sec2} presents the proposed taxonomy. Sections~\ref{sec3} through~\ref{sec6} systematically examine \del{each reasoning capability with corresponding tasks and methodological approaches}\add{the four capability domains with their corresponding methods and evaluations}. Section~\ref{sec7} \del{outlines promising future directions toward more capable physical AI systems}\add{then first analyzes positive and negative evidence for the proposed progression from perception to reasoning, modeling, and interaction, and subsequently discusses the remaining conceptual challenges for unified Physical AI}. Section~\ref{sec8} \del{synthesizes our contributions and findings}\add{concludes the survey}. 

\begin{table}[t]
\centering
\setlength{\tabcolsep}{5pt}
\renewcommand{\arraystretch}{1.05}
\caption{\del{Scope of existing surveys. They tend to examine only single dimensions of physical understanding abilities, without providing a comprehensive overview of physical AI evolution.}\add{Scope comparison of existing surveys across capability domains.}}
\label{tab:comparison}
\vspace{-4pt}
\begin{tabularx}{\linewidth}{
  >{\hsize=0.35\hsize\centering\arraybackslash}X
  >{\hsize=0.65\hsize\centering\arraybackslash}X
}
\toprule
\textbf{Capability} & \textbf{Existing Surveys} \\
\midrule
Perception & 
\cite{Yin2023ASO, Sapkota2025ObjectDW, Guo2024PhyGraspGR, Sun2025ProbingPC}\del{\cite{Marjieh2023LargeLM}} \\[-1pt]
Reasoning & 
\cite{Zhou2025FromPT, Li2025FromS1, Ravishankara2025TheAI, Duan2022ASO, Huang2022TowardsRI} \\[-1pt]
\multirow{2}{*}{Modeling} & 
\cite{Khan2025FoundationMD, Barman2025LargePM, Ding2024UnderstandingWO, Liu2025GenerativePA, Zhu2024IsSA, Kong20253DA4, Bergen2019MachineLF, Xie2025From2T, Chen2025ASO,Mai2024FromEM} \\[-1pt]
\multirow{2}{*}{Interaction}  & 
\cite{Liu2024AligningCS, Qi2024ShapeLLMU3, Sathyam2025FoundationMF, Feng2025EmbodiedAF, Long2025ASL, Wang2023ASO, Liang2025LargeME, Zheng2024ASO, Wang2025TowardEA,Duan2021ASOA,Xu2024ASOA,Han2023ASOA} \\
\bottomrule
\end{tabularx}
\vspace{-6pt}
\end{table}

%% file: sec/2_preliminaries.tex
\section{\del{Taxonomy}\add{Preliminaries}}
\label{sec2}

\del{To systematically understand how AI systems engage with physics, we propose a hierarchical taxonomy that reflects the natural progression from abstract reasoning to embodied action. Our framework organizes physics-aware AI capabilities into four interconnected domains that mirror human cognitive development in physical understanding. \textbf{Physical Perception} represents the ability to intuitively extract physical properties from sensory data through increasingly sophisticated layers, including object recognition, spatial relationships and intrinsic properties. In the inverse direction, \textbf{Physics Reasoning} encompasses symbolic manipulation and theoretical problem-solving, where AI systems leverage mathematical representations to solve physics problems ranging from textbook exercises to research-level challenges. These two complementary pathways converge in \textbf{World Modeling}, where AI systems integrate symbolic knowledge with perceptual understanding to construct predictive models of physical environments, enabling capabilities from video generation to reconstruction. Finally, \textbf{Embodied Interaction} grounds these cognitive capabilities in real-world action, where robots, autonomous vehicles, and navigation agents must reconcile theoretical understanding with the irreversible consequences of physical manipulation. Figure~\ref{fig:taxonomy} shows our taxonomy. This taxonomy not only delineates current research boundaries but also reveals critical gaps where symbolic reasoning and embodied experience remain disconnected, highlighting the path toward genuinely physics-aware artificial intelligence.}

\subsection{\add{Taxonomy}}
\input{sec/0TreeTaxonomy}

\add{To systematically characterize how AI systems engage with the physical world, we organize this survey using a capability--task hierarchy, shown in Figure~\ref{fig:taxonomy}. 
At the highest level, our framework organizes physics-aware AI capabilities into four interconnected domains that mirror human cognitive development in physical understanding. It groups Physical AI systems according to the principal type of output they produce when operating on physical environments. Under this view, \textbf{Physical Perception} estimates the current hidden physical state from observations, \textbf{Physics Reasoning} produces symbolic answers or derivations, \textbf{World Modeling} predicts future states or observations under physical dynamics, and \textbf{Embodied Interaction} outputs actions that intervene in the environment. This output-centric perspective makes the taxonomy applicable across different model families, ranging from classical structured systems to modern foundation models.}

\add{At the second level, each capability is decomposed into task families. This hierarchical design reflects the progression from passive observation to active intervention. Under \textbf{Physical Perception}, \del{tasks progress from scene and object understanding to spatial perception, intrinsic property estimation, dynamic state estimation, and causal structure inference}\add{the task families are object recognition, spatial perception, intrinsic property estimation, and dynamic estimation}. Under \textbf{Physics Reasoning}, \del{the focus shifts from benchmark-oriented problem solving to general physical reasoning and theoretical or solver-based settings}\add{the task families are symbolic reasoning, multimodal-grounded reasoning, causal and counterfactual reasoning, and accelerate physics research}. Under \textbf{World Modeling}, \del{the emphasis moves to predictive simulation, generative modeling, and physics-enhanced modeling fidelity}\add{the task families are image generation, video generation, scene reconstruction, and physics-constrained simulation}. Finally, under \textbf{Embodied Interaction}, the task families are organized around robotics, navigation, and autonomous driving, where the system must couple physical understanding with real-time decision making.}

\add{Importantly, this hierarchy is intended to capture what a model primarily does, rather than how it is implemented. Accordingly, architectural choices (e.g., transformer, diffusion model, GNN, differentiable simulator), benchmark suites, and application domains are treated as auxiliary descriptors rather than primary taxonomy axes. This separation improves interpretability and avoids conflating task objectives with model families or evaluation settings.}

\subsection{\add{Formal Definitions}}

\add{To delimit the scope of this survey, we formalize Physical AI within a partially observed controlled dynamical system.}
\add{At time step $t$, let $s_t \in \mathcal{S}$ denote the latent physical state of the environment, $o_t \in \mathcal{O}$ the observation and $a_t \in \mathcal{A}$ the action.}
\add{We define Physical AI as a partially observed controlled dynamical system:}
\begin{equation}
\add{s_{t+1} \sim p_{\mathrm{env}}(s_{t+1}\mid s_t,a_t),}
\quad
\add{o_t \sim p_{\mathrm{env}}(o_t\mid s_t),}
\label{eq:env_dynamics}
\end{equation}
\add{We denote the interaction history by}
\begin{equation}
\add{h_t := (o_{1:t}, a_{1:t-1}),}
\label{eq:history}
\end{equation}
\add{and use it as the observable context under partial observability.}

\begin{table}[t]
\centering
\caption{\add{Unified notation used in our formalization.}}
\label{tab:notation}
\setlength{\tabcolsep}{5pt}
\renewcommand{\arraystretch}{1.08}
\begin{tabularx}{\linewidth}{>{\raggedright\arraybackslash}p{0.22\linewidth}X}
\toprule
\add{\textbf{Symbol}} & \add{\textbf{Meaning}} \\
\midrule
\add{$s_t \in \mathcal{S}$} & \add{Latent physical state of environment in $t$} \\
\add{$o_t \in \mathcal{O}$} & \add{Observation/partial or noisy view of $s_t$} \\
\add{$a_t \in \mathcal{A}$} & \add{Action or executable intervention of agent in $t$} \\
\add{$l \in \mathcal{L}$} & \add{Language instruction} \\
\add{$q \in \mathcal{Q}$} & \add{Symbolic physics query} \\
\add{$h_t$} & \add{Interaction history} \\
\add{$\kappa \in \mathcal{K}$} & \add{Task-relevant physical law or constraint set} \\
\add{$r_t$} & \add{Optional reasoning trace} \\
\add{$x_t = g(s_t)$} & \add{Task-relevant physical state, property, or structure} \\
\add{$y_t^{\mathrm{R}}$} & \add{Task-level reasoning output} \\
\bottomrule
\end{tabularx}
\end{table}

\add{\noindent\textbf{Physical Perception.}}
\add{Physical Perception concerns inferring task-relevant physical state, property, or structure from observations.}
\add{Let $x_t = g(s_t)$ denote the task-relevant readout of the underlying physical state, where $g(\cdot)$ may extract geometry, material, depth, contact relations, motion cues, or other structured descriptors.}
\add{Then physical perception is defined as}
\begin{equation}
\add{p_\theta(x_t \mid h_t).}
\label{eq:perception_dist}
\end{equation}
\begin{equation}
\add{\hat{x}_t = \argmax_x p_\theta(x \mid h_t).}
\label{eq:perception}
\end{equation}
\add{Here, $\hat{x}_t$ denotes the model-estimated task-relevant physical variable at time step $t$. Its output is a present-time estimate of physical state or structure.}
\add{This category includes static property estimation and short-horizon dynamic inference so long as the method's primary role is to recover physically meaningful variables from observations, rather than to build a reusable simulator or directly choose actions.}

\add{\noindent\textbf{Physics Reasoning.}}
\add{Physics Reasoning concerns applying, composing, or discovering physical laws and constraints to solve a physics task.}
\add{Given a physics query $q \in \mathcal{Q}$ and relevant laws $\kappa \in \mathcal{K}$, it is defined by}
\begin{equation}
\add{p_\theta(y_t^{\mathrm{R}}, r_t, \kappa \mid h_t, q).}
\label{eq:reasoning_joint}
\end{equation}
\begin{equation}
\add{(\hat{y}_t^{\mathrm{R}}, \hat{r}_t, \hat{\kappa}) = \argmax_{y,r,\kappa} p_\theta(y, r, \kappa \mid h_t, q).}
\label{eq:reasoning}
\end{equation}
\add{Here $y_t^{\mathrm{R}}$ denotes the task-level reasoning output, which may take the form of an answer, derivation, explanation, numerical solution, discovered law, or executable scientific artifact, and $r_t$ is an optional reasoning trace. $\hat{\kappa}$ is the physical law or constraint set selected, composed, or discovered by the model.}
\add{Under this definition, physics reasoning includes symbolic derivation, multimodal-grounded explanation, causal or counterfactual inference, and research-acceleration workflows (e.g., hypothesis generation and streamlined experimentation), provided that the central competence is the use or discovery of physical principles.}

\add{\noindent\textbf{World Modeling.}}
\add{World Modeling concerns constructing a physically coherent model of the environment that supports generation, reconstruction, prediction, or simulation.}
\add{Using hats to denote model-generated states or observations, we define the current-world form as}
\begin{equation}
\add{p_\theta(\hat{s}_t, \hat{o}_t \mid h_t, l).}
\label{eq:wm_onestep}
\end{equation}
\add{The rollout form is}
\begin{equation}
\add{p_\theta(\hat{s}_{t+1:t+H}, \hat{o}_{t+1:t+H} \mid h_t, a_{t:t+H-1}, l).}
\label{eq:wm_rollout}
\end{equation}
\add{The first form covers generation or reconstruction of the current world state, while the second covers future prediction and simulation.}
\add{This capability therefore includes image generation, video generation, scene reconstruction, and physics-constrained simulation.}
\add{The key distinction from perception is that world modeling aims to build a reusable generative or simulatable representation of the environment, rather than only estimating task-specific physical variables.}

\add{\noindent\textbf{Embodied Interaction.}}
\add{Embodied Interaction concerns selecting executable interventions in a physical environment.}
\add{Under full observability, it is defined as}
\begin{equation}
\add{\pi_\theta(a_t \mid s_t, l).}
\label{eq:interaction_full}
\end{equation}
\add{Under partial observability, it becomes}
\begin{equation}
\add{\pi_\theta(a_t \mid h_t, l).}
\label{eq:interaction_partial}
\end{equation}
\add{Here $a_t$ may denote low-level controls or immediately executable action parameterizations such as waypoints or planned trajectories.}
\add{Thus, embodied interaction covers both direct control policies and decision-making modules whose outputs are deployed in a closed loop to change the physical environment.}


\begin{table*}[t]
\centering
\caption{\add{Feature matrix for the four top-level capabilities in our taxonomy. The classification is determined by the principal physical function and dominant evaluation role of a method, rather than by a single narrowly defined output variable, internal modules, or training strategy.}}
\label{tab:feature_matrix}
\setlength{\tabcolsep}{4pt}
\renewcommand{\arraystretch}{1.12}
\begin{tabularx}{\textwidth}{
>{\raggedright\arraybackslash}p{0.10\textwidth}
>{\raggedright\arraybackslash}p{0.10\textwidth}
>{\raggedright\arraybackslash}p{0.15\textwidth}
>{\centering\arraybackslash}p{0.10\textwidth}
>{\raggedright\arraybackslash}p{0.10\textwidth}
X}
\toprule
\add{\textbf{Capability}} & \add{\textbf{Output}} & \add{\textbf{Temporal Scope}} & \add{\textbf{Executable}} & \add{\textbf{Reasoning}} & \add{\textbf{Typical Metrics}} \\
\midrule
\add{Perception} 
& \add{$\hat{x}_t$} 
& \add{Present} 
& \add{$\times$} 
& \add{Low} 
& \add{Detection / depth / attribute error} \\

\add{Reasoning} 
& \add{$(\hat{y}^{\mathrm{R}}, \hat{r})$} 
& \add{Present / timeless} 
& \add{$\times$} 
& \add{High} 
& \add{Answer accuracy / derivation correctness} \\

\add{Modeling} 
& \add{$(\hat{s}, \hat{o})$} 
& \add{Future} 
& \add{$\times$} 
& \add{Optional} 
& \add{Rollout error / FVD / consistency} \\

\add{Interaction} 
& \add{$\hat{a}_t$} 
& \add{Online} 
& \add{$\checkmark$} 
& \add{Optional} 
& \add{Success / return / safety} \\
\bottomrule
\end{tabularx}
\end{table*}

%% file: sec/0TreeTaxonomy.tex

\begin{figure*}[t]
\centering

\definecolor{userRoot}{HTML}{225ea8}
\definecolor{userL1}{HTML}{41b6c4}
\definecolor{userL2}{HTML}{a1dab4}
\definecolor{userL3}{HTML}{ffffcc}

\begin{tikzpicture}[
  node distance=0.15cm,
  every node/.style={
    draw,
    rectangle,
    rounded corners=2pt,
    minimum height=0.5cm,
    text centered,
    font=\tiny,
    inner sep=1pt
  },
  Root/.style={
    fill=userRoot, text=white,
    minimum width=8cm,
    font=\footnotesize\bfseries
  },
  Level1/.style={
    fill=userL1, text=white,
    minimum width=3.2cm,
    font=\scriptsize,
    text width=3.0cm
  },
  Level2/.style={
    fill=userL2,
    minimum width=3.0cm,
    text width=2.8cm
  },
  Level3/.style={
    fill=userL3,
    minimum width=2.8cm,
    text width=2.6cm
  }
]


\node[Root] (root) at (0,0) {Physical AI Systems};

\node[Level1] (physical) at (-5.5, -1.5) {Physical Perception};
\node[Level1] (physics) at (-1.8, -1.5) {Physics Reasoning};
\node[Level1] (world) at (1.8, -1.5) {World Modeling};
\node[Level1] (embodied) at (5.5, -1.5) {Embodied Interaction};

\draw[->] (root.south) -- ++(0,-0.4) -| (physical.north);
\draw[->] (root.south) -- ++(0,-0.4) -| (physics.north);
\draw[->] (root.south) -- ++(0,-0.4) -| (world.north);
\draw[->] (root.south) -- ++(0,-0.4) -| (embodied.north);

\node[Level2, below=0.1cm of physical] (object_rec) {Object Recognition};
\node[Level3, below=0.05cm of object_rec] (category_scene_rec) {Category \& Scene Recognition};
\node[Level3, below=0.05cm of category_scene_rec] (anomaly_detection) {Anomaly Detection};
\node[Level2, below=0.1cm of anomaly_detection] (spatial_perc) {Spatial Perception};
\node[Level3, below=0.05cm of spatial_perc] (spatial_relationships) {2D \& 3D Relations};
\node[Level3, below=0.05cm of spatial_relationships] (metric_semantic_mapping) {Metric-Semantic Mapping};
\node[Level2, below=0.1cm of metric_semantic_mapping] (intrinsic_prop) {Intrinsic Property Estimation};
\node[Level3, below=0.05cm of intrinsic_prop] (material_optical) {Material \& Optical Properties};
\node[Level3, below=0.05cm of material_optical] (mass_rigidity) {Mass \& Rigidity Estimation};
\node[Level2, below=0.1cm of mass_rigidity] (dynamic_est) {Dynamic Estimation};
\node[Level3, below=0.05cm of dynamic_est] (interaction_dynamics) {Interaction Dynamics};
\node[Level3, below=0.05cm of interaction_dynamics] (state_prediction) {State Prediction};

\node[Level2, below=0.1cm of physics] (symbolic_reasoning) {Symbolic Reasoning};
\node[Level3, below=0.05cm of symbolic_reasoning] (textbook_exam) {Textbook \& Exam QA};
\node[Level3, below=0.05cm of textbook_exam] (olympiad_reasoning) {Olympiad Problem Solving};
\node[Level2, below=0.1cm of olympiad_reasoning] (multimodal_reasoning) {Multimodal-grounded Reasoning};
\node[Level3, below=0.05cm of multimodal_reasoning] (diagram_reasoning) {Diagram-grounded Reasoning};
\node[Level3, below=0.05cm of diagram_reasoning] (video_reasoning) {Video-grounded Reasoning};
\node[Level2, below=0.1cm of video_reasoning] (causal_reasoning) {Causal \& Counterfactual Reasoning};
\node[Level3, below=0.05cm of causal_reasoning] (causal_discovery) {Causal Discovery};
\node[Level3, below=0.05cm of causal_discovery] (intervention_analysis) {Intervention Analysis};
\node[Level2, below=0.1cm of intervention_analysis] (research_automation) {Accelerate Physics Research};
\node[Level3, below=0.05cm of research_automation] (hypothesis_generation) {Hypothesis Generation};
\node[Level3, below=0.05cm of hypothesis_generation] (tool_simulation) {Streamline Experimentation};

\node[Level2, below=0.1cm of world] (image_gen) {Image Generation};
\node[Level3, below=0.05cm of image_gen] (material_rendering) {Material-aware Rendering};
\node[Level3, below=0.05cm of material_rendering] (geometry_synthesis) {Geometry-consistent Synthesis};
\node[Level2, below=0.1cm of geometry_synthesis] (video_gen) {Video Generation};
\node[Level3, below=0.05cm of video_gen] (future_frames) {Future Frame Prediction};
\node[Level3, below=0.05cm of future_frames] (physics_video) {Physics-consistent Video Synthesis};
\node[Level2, below=0.1cm of physics_video] (scene_recon) {Scene Reconstruction};
\node[Level3, below=0.05cm of scene_recon] (recon_3d) {3D Scene Recovery};
\node[Level3, below=0.05cm of recon_3d] (recon_4d) {4D Dynamic Reconstruction};
\node[Level2, below=0.1cm of recon_4d] (phys_sim) {Physics-constrained Simulation};
\node[Level3, below=0.05cm of phys_sim] (counterfactual_rollout) {Counterfactual Rollouts};
\node[Level3, below=0.05cm of counterfactual_rollout] (intervention_sim) {Intervention-based Simulation};

\node[Level2, below=0.1cm of embodied] (robotics) {Robotics};
\node[Level3, below=0.05cm of robotics] (cont_action) {Continuous Action Generation};
\node[Level3, below=0.05cm of cont_action] (cross_platform) {Cross-platform Generalization};
\node[Level2, below=0.1cm of cross_platform] (navigation) {Navigation};
\node[Level3, below=0.05cm of navigation] (object_nav) {Object-Goal Navigation};
\node[Level3, below=0.05cm of object_nav] (vln_nav) {Vision-Language Navigation};
\node[Level3, below=0.05cm of vln_nav] (dialog_nav) {Dialog-based Navigation};
\node[Level2, below=0.1cm of dialog_nav] (auto_driving) {Autonomous Driving};
\node[Level3, below=0.05cm of auto_driving] (planning_policies) {Planning-oriented Policies};
\node[Level3, below=0.05cm of planning_policies] (sim_assisted) {Simulation-assisted Driving};

\end{tikzpicture}
\caption{The proposed capability--task--subtask taxonomy of Physical AI. The hierarchy is organized by the principal output of each capability, while benchmarks, architectures, and training strategies are discussed in the text rather than treated as taxonomy nodes. \add{The teal boxes denote the four capability-level, output-centric groups discussed in Sections~\ref{sec3}--\ref{sec6}; the green boxes denote task families corresponding to subsections within those capability sections; and the pale-yellow boxes denote representative subtasks discussed under the corresponding task families.}}
\label{fig:taxonomy}
\end{figure*}

%% file: sec/3_physical_reasoning.tex
\section{Physical Perception: From Sensory Data to Physical Understanding}
\label{sec3}

Understanding the physical world begins with perception, where AI systems transform raw sensory inputs into structured representations of physical properties, spatial relationships, dynamic behaviors. Physical perception serves as the essential grounding for subsequent reasoning, modeling, embodied interaction, establishing the observational foundation upon which symbolic reasoning and predictive models are built. While traditional computer vision focuses on semantic recognition, physical perception demands extracting geometry, material properties, force dynamics, causal structure from visual or multimodal observations. This capability enables systems to infer latent physical states that are not directly observable but fundamentally govern real-world phenomena. We organize existing investigations along a progression of increasing cognitive complexity: Object Recognition extracts geometric structure with semantic information, Spatial Perception captures relationships within scene composition, Intrinsic Property Estimation infers material characteristics and Dynamic Estimation predicts motion through temporal evolution\del{, Causal and Counterfactual Inference reveals underlying physical mechanisms}. This hierarchical organization reflects how perceptual capabilities build cumulatively, with each level providing foundations for the next, ultimately enabling the transition from observation to reasoning about physical laws.

\subsection{Object Recognition}
\add{\noindent\textbf{Methods.}}
The most fundamental aspect of visual physical perception is the ability to identify objects and determine their spatial relationships within a given scene.
In the past decade, the development of convolutional neural networks (CNNs) has made it possible to solve target detection and object classification problems. \webref[GPT-4V system card]{GPT-4V}{https://openai.com/index/gpt-4v-system-card/}{Apr. 13, 2026} serves as a milestone that demonstrates robust zero-shot object detection and localization capabilities of MLLMs across diverse visual contexts.
Other open-sourced models have also revealed that the introduction of high-quality labeled data can help agents recognize objects at multiple levels of granularity, from basic categories (e.g., "vehicle," "animal") to fine-grained classifications (e.g., "Maserati," "golden retriever")~\cite{ he2018maskrcnn,liu2024groundingdinomarryingdino,xu2025qwen3omnitechnicalreport}. For more complex scene-level recognition tasks, MLLMs should integrate individual object detections into a coherent understanding of the environment. This involves recognizing scene categories (indoor/outdoor, kitchen/bedroom), understanding typical object arrangements, and identifying anomalous configurations~\cite{Song_2024,varghese2025viewinvariantpixelwiseanomalydetection,zhang2024cognition,Ji_2024}. \del{Recent evaluations}\del{~\cite{MME,Roboflow100-VL,EagleVision,ROD-MLLM}}\del{ have shown that MLLMs have achieved competitive performance on object-level detection capabilities, which lays the foundation for subsequent physical property perception works.}

\add{\noindent\textbf{Evaluation Landscape.} Existing benchmarks for physical perception range from generic recognition to physically grounded evaluation.}
\add{While classic benchmarks such as ImageNet~\cite{Imagenet} and COCO~\cite{coco} mainly assess category recognition and detection, they do not explicitly test pose estimation or physically abnormal states.}
\add{BOP~\cite{BOP} provides a more relevant setting for 6D object localization, where the top method in the 2020 challenge achieves 69.8 AR, indicating that pose-aware recognition is feasible but still sensitive to viewpoint and occlusion.}
\add{For anomaly perception, Phys-AD~\cite{Phys-AD} contains 6,359 videos across 22 categories and 47 anomaly types; strong unsupervised baselines achieve only around 52\% AUROC on average, highlighting the difficulty of recognizing physical anomalies beyond appearance cues~\cite{Phys-AD}.}

\subsection{Spatial Perception}
\add{\noindent\textbf{Methods.}}
Beyond object recognition, AI systems must understand spatial relationships to build coherent scene representations. This includes both absolute positioning (e.g., "at the center of the whole image") and relative positioning (e.g., "to the left of the white table").
\del{Many recent benchmarks are specifically designed to test spatial understanding, such as VSR and SpatialBench~\cite{SpatialScore,Open3DVQA,yang2025mmsi,wu2025spatialmllmboostingmllmcapabilities,xu2025multi}, reveal varying capabilities across different MLLMs.} \add{Recent methods explore spatially aware representation learning and multimodal grounding~\cite{SpIRL,wu2025spatialmllmboostingmllmcapabilities,xu2025multi}, together with multi-view formulations that explicitly enforce cross-view correspondence~\cite{ViewSpatial,zheng2023comprehensive}, revealing varying capabilities across different MLLMs.}
Models generally perform well on basic 2D spatial prepositions (above, below, left, right) but struggle with more complex spatial concepts such as 3D spatial reasoning, pixel-level localization and scale relationships. The ability to handle these complex spatial tasks is typically limited by the nature of their training data and architectures\del{. Most models rely on large-scale datasets like COCO}\del{~\cite{coco}}\del{ or ImageNet}\del{~\cite{Imagenet}}\del{, which focus on object identification and simple relationships but lack detailed annotations for more sophisticated spatial reasoning tasks.}\add{, since most current MLLMs are optimized primarily for semantic alignment and therefore often exhibit weak geometric grounding when precise spatial relations must be resolved.}
\add{These methods mainly improve spatial perception in two ways: first, by introducing explicit spatial inductive biases to preserve relative layout and local geometric structure; second, by exploiting cross-view correspondence and multi-frame consistency to recover 3D relations that are ambiguous in single-view inputs.}

\add{\noindent\textbf{Evaluation Landscape.} }\add{Spatial perception is increasingly evaluated by dedicated benchmarks such as SpatialScore~\cite{SpatialScore}, Open3DVQA~\cite{Open3DVQA}, and MMSI-Bench~\cite{yang2025mmsi}, which probe 2D/3D relations, metric grounding, and cross-view consistency.} \add{Open3DVQA shows that current MLLMs are more reliable on relative relations than absolute ones~\cite{Open3DVQA}.} \add{MMSI-Bench reports that the strongest open-source model reaches only about 30\% accuracy and the best proprietary model about 40\%, far below 97\% human accuracy~\cite{yang2025mmsi}.} \add{These results suggest that precise geometric grounding, rather than coarse relational recognition, remains the central challenge.}

\subsection{Identifying Intrinsic Property}
\add{\noindent\textbf{Methods.}}
Understanding the physical world from vision requires not only recognizing objects but also inferring their intrinsic properties and dynamic behaviors based on these properties.
Intrinsic properties such as mass, viscosity and rigidity are inherent characteristics of objects that remain constant regardless of observational perspectives.
Estimating these properties from visual observation alone is particularly challenging for AI as it demands mapping visual features to physical attributes that may not be directly observable but can only be inferred by the laws of physics.
Recent studies aim to achieve reliable identification of material such as metal, fabric and plastic~\cite{kocsis2024intrinsicimagediffusionindoor}\del{\cite{Van_Zuijlen_2021}}\del{ and even finer textures (e.g., silk vs. cotton)}\add{, finer textures}\del{~\cite{jang2024fabric}}~\cite{hu2020fabricsurfacecharacterizationassessment}\del{, while also identifying optical properties such as transparency and translucency}\add{ while also identifying optical properties such as transparency and translucency}\del{~\cite{weidenbach2024transparency,nagai2025top}}~\cite{chen2024practical}.
For mass \del{and weight} estimation, models \del{primarily rely on size cues, material-density associations, and category priors, showing stronger performance in relative rather than absolute weight prediction, though systematic biases persist}\add{typically combine visual size cues, category priors, and learned appearance--weight correlations, and are usually more reliable in restricted domains than in open-world settings}~\cite{Vision-Based,M_ller_2024}\del{\cite{afridi2024analyzing,nath2024mass}}.
In terms of rigidity\del{ and deformability, MLLMs can classify rigid and flexible objects, detect evidence of deformation, and exploit contextual cues such as force interactions, but their judgments remain heuristic and fragile in complex settings}\add{, some methods infer plausible deformation or interaction outcomes from static visual evidence, but judgments remain fragile in complex settings}\del{~\cite{mustafa2025forcemapping,song2025moda,PaNDaS}}~\cite{PhysID}.
\del{Nevertheless, correctly identifying ambiguous cases, novel materials, and lighting variability remain major challenges for frontier models.}\add{Nevertheless, ambiguous materials, unseen objects, and lighting variation still make intrinsic-property estimation much less stable than ordinary semantic perception.}

\add{\noindent\textbf{Evaluation Landscape.} }\add{Evaluation in this task family remains fragmented.} \del{MIP~\cite{Van_Zuijlen_2021}, with 19,325 paintings and over 200K material bounding boxes across 15 coarse and 50+ fine-grained categories, can be used to benchmark material perception.} \add{Material and texture estimation are still evaluated mainly through task-specific datasets and narrow experimental settings rather than through a single widely adopted benchmark~\cite{kocsis2024intrinsicimagediffusionindoor,hu2020fabricsurfacecharacterizationassessment}.} \add{By contrast, mass estimation still depends on domain-specific datasets, such as the food-weight dataset in~\cite{Vision-Based} containing 2,380 images over 14 food types.} \add{Overall, intrinsic-property perception lacks a unified benchmark comparable to object or spatial perception.}

\subsection{Dynamic Estimation}
\add{\noindent\textbf{Methods.}}
Crucially, these intrinsic properties serve as foundation for understanding how objects behave according to the laws of physics. Knowledge of mass and rigidity, for instance, directly informs whether/how an object might deform under external force, or how it influences motion during collisions. Building on this static perspective, dynamic property perception captures how objects behave and interact over time through contact, constraints, and forces such as support, occlusion, friction, and impact.
Unlike intrinsic estimation that answers what an object is in terms of characteristics, dynamic perception addresses how it behaves upon interaction with other objects.
\add{Graph-based methods, especially graph neural networks (GNNs), have been central to early studies of dynamic perception.}
\add{Models such as Interaction Networks, Visual Interaction Networks, and the Neural Physics Engine~\cite{interactionnetworks,VIN,NPE} infer object relations from observations and expose contact or interaction structure, although they already lie near the boundary between perception and world modeling because they also support short-horizon rollouts.}
\add{More recent work extends this line in two directions: inferring likely dynamics from weaker visual evidence, such as static images or raw videos~\cite{NewtonianIU,PhysID}, and using object-centric video models as strong auxiliary baselines when the goal is to preserve interaction structure over time~\cite{SlotFormer,li2025reasoningenhancedobjectcentriclearningvideos}.}
\add{Together, these methods move dynamic estimation from simple motion extrapolation toward richer perception of contact, support, and physically plausible state change.}
\del{More recent frameworks like I-PHYRE}\del{~\cite{I-PHYRE}}\del{ further challenge agents to exhibit intuitive physical reasoning, multi-step planning, and in-situ intervention tasks, highlighting the importance of real-time dynamic perception.}\del{ In addition, benchmarks such as DeepPHY}\del{~\cite{DeepPHY}}\del{ provide systematic evaluation environments that couple intrinsic attributes with dynamic outcomes, ensuring models are tested on both the static and temporal dimensions of physical understanding.}
\add{\noindent\textbf{Evaluation Landscape.} }\add{Dynamic estimation is commonly evaluated by Physion~\cite{Physion}, Physion++~\cite{PhysionPlus}, I-PHYRE~\cite{I-PHYRE}, and ContPhy~\cite{ContPhy}.} \add{Physion++ is particularly informative because it requires online inference of latent physical properties from videos: under the main separate-training setting, ALOE reaches 53.4\% overall accuracy and SlotFormer 56.7\%, while human accuracy is about 60\%; the best model--human correlation is only $r=0.12$, compared with a split-half human correlation of $r=0.37$~\cite{PhysionPlus}.} \add{I-PHYRE and ContPhy further show that performance degrades when evaluation shifts from passive prediction to intervention-heavy or continuum-dynamics settings~\cite{I-PHYRE,ContPhy}.}
By moving from the recognition of intrinsic object properties to the perception of their dynamic relations, AI systems develop a more comprehensive human-like understanding of physical environments.

%% file: sec/4_physics_reasoning.tex
\section{Physics Reasoning: From Observation to Explanation}
\label{sec4}

Physics reasoning represents the critical transition from perception to explanation, where systems move beyond identifying physical properties to understanding why phenomena occur. Building directly upon perceptual foundations, reasoning operates through equations and theoretical principles that encode physical laws. While perception answers "what is happening" by extracting observable properties, reasoning addresses "why it happens" by connecting abstract concepts with concrete phenomena through causal relationships governed by \del{natural laws}\add{laws of nature}. This capability bridges observational understanding with predictive modeling, enabling systems to not merely recognize patterns but derive explanations grounded in physical principles. 
We examine how AI systems perform physics reasoning across \del{two broad domains of increasing sophistication}\add{four task families: symbolic reasoning, multimodal-grounded reasoning, causal and counterfactual reasoning, and accelerate physics research}1g. This progression reflects the deepening cognitive demands from applying established principles to discovering novel theoretical insights, ultimately preparing systems for the modeling stage where learned physical laws enable forward prediction and counterfactual simulation.

\begin{figure*}[th]
    \centering
    \includegraphics[width=1\textwidth]{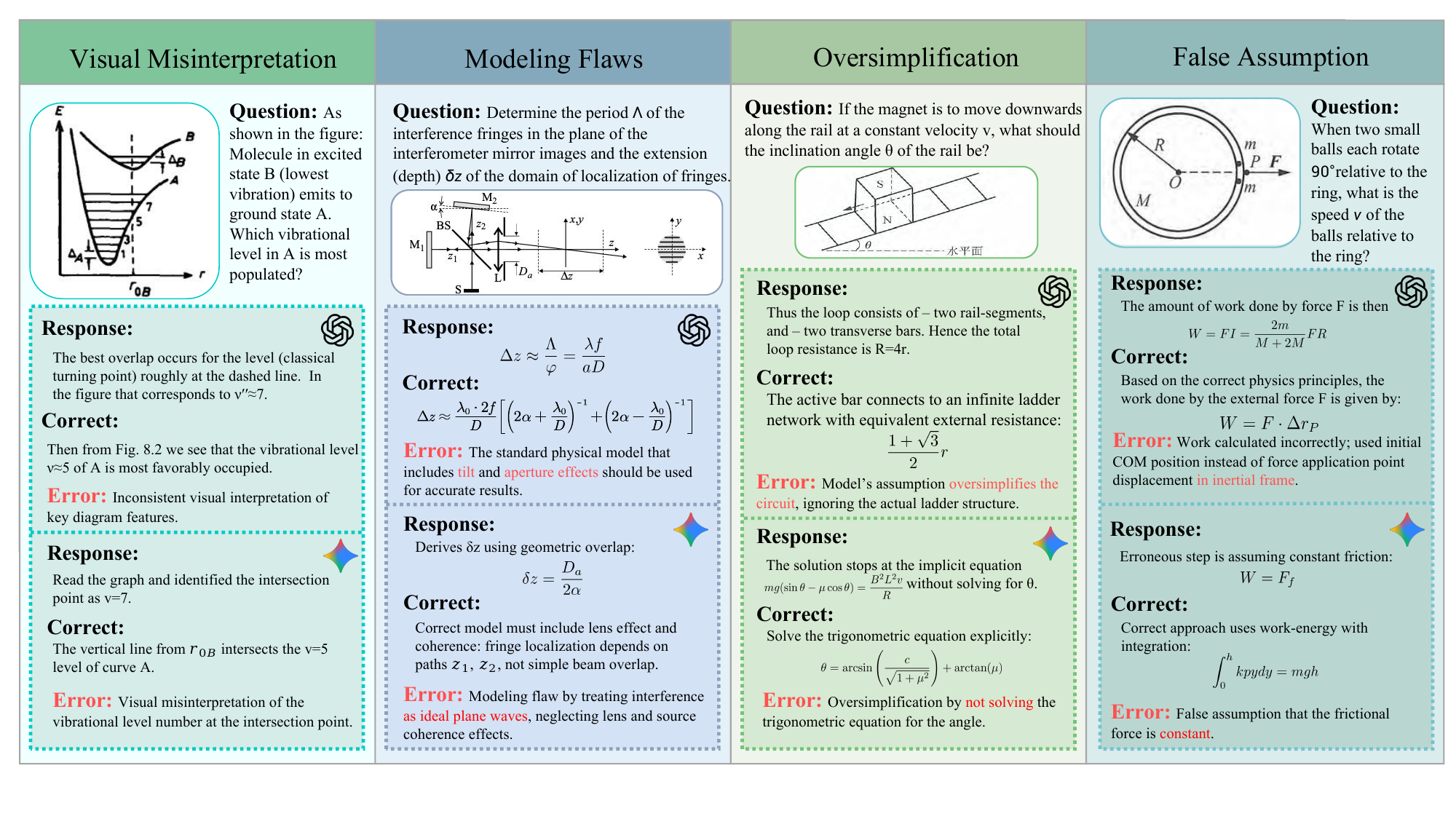} 
    \caption{Error pattern analysis of closed-Source models in multimodal physical reasoning.}
    \label{fig:physics_case}
\end{figure*}

\begin{table}[t]
\centering
\caption{Performance (\%) of  LLMs on representative text-based and multimodal physics benchmarks. GPT: GPT-4o~\cite{gpt4o}, Claude: Claude-3.7-Sonnet, DeepSeek: DeepSeek-R1~\cite{guo2025deepseek}, Gemini: Gemini-2.5-pro.}
\label{tab:llm_bmk}

\begin{tabularx}{\linewidth}{>{\raggedright\arraybackslash}l
                             >{\centering\arraybackslash\hsize=0.6\hsize}X
                             >{\centering\arraybackslash\hsize=1.0\hsize}X
                             >{\centering\arraybackslash\hsize=1.3\hsize}X
                             >{\centering\arraybackslash\hsize=1.1\hsize}X}
\toprule
\textbf{Benchmark} & \textbf{GPT} & \textbf{Claude} & \textbf{DeepSeek} & \textbf{Gemini} \\
\midrule
\multicolumn{5}{l}{\textbf{Text-based}} \\ 
UGPhysics~\cite{ugphysics}           & 38.66 & -- & 56.34 & -- \\
PHYBench~\cite{qiu2025phybench}      & 6.97 & 12.87 & 33.45 & 45.82  \\
GPQA~\cite{GPQA}               & 50.3 & 67.4 & 71.5 & 84.0  \\
OlympiadBench~\cite{olympiadbench}       & 39.72 & -- & -- & 7.34  \\
\midrule
\multicolumn{5}{l}{\textbf{Multimodal}} \\  
SeePhys~\cite{seephys}            & 21.9 & 34.6 & 42.2 & 54.9  \\
PhysReason~\cite{physreason}  & 29.58 & -- & 34.07 & -- \\
MMMU~\cite{mmmu}        & 59.4 & 75.0 & -- & 81.7 \\
MMMU-Pro~\cite{mmmupro}    & 51.9 & 76.4 & -- & -- \\
\bottomrule
\end{tabularx}

\end{table}

\del{\noindent\textbf{Benchmarking Physics Reasoning}}
\del{We first systematically review the current landscape of physics reasoning benchmarks from both text-based and multimodal perspectives, shown in Table~\ref{tab:llm_bmk}.}
\del{\noindent\textbf{Text-based Benchmarks. }}
\del{Early benchmarks primarily focused on textbook physics problems in pure textual format, establishing baselines for symbolic manipulation and numerical reasoning.}
\del{Representative resources such as PhysicsEval}\del{~\cite{physicseval}}\del{ compile problems across multiple education levels and emphasize inference-time techniques, numerical calculation, and robustness under distributional shifts.}
\del{These benchmarks provide systematic coverage from fundamental mechanics and kinematics to electricity, thermodynamics, and modern physics, enabling evaluation of both single-step calculations and multi-step reasoning chains.}
\del{Other text-based datasets, including UGPhysics}\del{~\cite{ugphysics}}\del{, PHYBench}\del{~\cite{qiu2025phybench}}\del{, GPQA}\del{~\cite{GPQA}}\del{, further extend the scope of questions to university-level courses and examinations.}
\del{Empirical studies consistently show that even state-of-the-art language models often depend on superficial pattern matching and memorization, with sharp performance degradation on tasks requiring genuine physical reasoning, long multi-step inference, or generalization beyond the training distribution.}
\del{Such findings highlight the persistent gap between current models and the robust reasoning capabilities expected from human-level or world-model-based physical understanding, motivating future benchmarks that more explicitly target multi-step reasoning, causal inference, and compositional problem-solving.}
\del{\noindent\textbf{Multimodal Benchmarks. }}
\del{On another dimension, a fundamental distinction between physics reasoning and pure mathematics lies in the involvement of more complex visual perception and diagram comprehension, which has driven the development of multimodal benchmarks.}
\del{Typical multimodel benchmarks like SeePhys}\del{~\cite{seephys}}\del{ highlight the importance of diagram-dependent physical reasoning by organizing problems across seven domains and twenty-one categories, demonstrating that visual context is often indispensable for accurate solutions.}
\del{Building on this line, PhysReason}\del{~\cite{physreason}}\del{ expands the paradigm by introducing multimodal diagnostic tasks and stepwise scoring to assess models' ability to integrate perceptual cues with symbolic reasoning.}
\del{More recent benchmarks, including ScienceQA}\del{~\cite{scienceqa}}\del{ for science-oriented diagram comprehension, MME-SCI and MMMU series}\del{~\cite{mme_sci,mmmu,mmmupro}}\del{ for large-scale perception--reasoning evaluation, and SciEval}\del{~\cite{scieval}}\del{ for competition-level and interdisciplinary multimodal science problems, further broden the evaluation landscape by emphasizing visual perception, causal inference, and counterfactual reasoning under dynamically varied conditions.}
\del{Together, these benchmarks extend the evaluation focus from purely text-based problem solving to scenarios where perception and reasoning must jointly operate in complex visual environments.}
\del{As shown in Figure~\ref{fig:physics_case}, performance analyses on these multimodal benchmarks consistently reveal that current models excel at pattern recognition but struggle to connect perceptual input with underlying physical principles, especially in long-horizon or distribution-shifted scenarios.}
\del{This gap underscores that progress in physical reasoning will require not only more sophisticated perception modules but also architectures capable of structured causal reasoning and compositional inference.}
\del{Developing benchmarks with richer multimodal tasks and more realistic dynamic scenes will be crucial for promoting models capable of integrating perception, reasoning, and simulation into a coherent world-model of physical systems.}

\subsection{\add{Symbolic Reasoning}}
\add{\noindent\textbf{Methods.}}
\add{Symbolic reasoning focuses on physics problems whose solution is explicitly expressed as a numerical answer, analytical derivation, equation, or verbal explanation.}
\add{Typical tasks include textbook problem solving, exam-style question answering, and olympiad-level deduction.}
\add{Unlike physical perception, the objective is not to estimate scene attributes such as depth, contact, or material, but to infer what follows from the observed or described conditions under explicit physical constraints, such as force balance, conservation laws, constitutive relations, or boundary conditions.}

\add{Recent progress in this area has moved from early Chain-of-Thought prompting methods~\cite{CoT} toward reinforcement learning and multi-agent paradigms.}
\add{Physics Supernova~\cite{PhysicsSupernova} demonstrates that tool-augmented AI agents can approach elite human performance on olympiad-level physics tasks, while LOCA-R~\cite{Jian2025LOCA} formulates Chinese Physics Olympiad solving as a sequence of localized and verifiable reasoning steps, substantially improving long-horizon solution reliability.}
\add{Prompt-based methods~\cite{Kortemeyer2023CouldAAF,Addala2024StepsAAA,Dan2025SymbolicOND} show that LLMs can solve a non-trivial subset of textbook physics problems, but their performance remains brittle under long multi-step derivations and distribution shift.}
\add{More recent methods therefore introduce explicit scaffolds for law-based inference.}
\add{Addala et al.~\cite{Addala2024KnowledgeGAE} show that external knowledge graphs can help decompose and constrain physics question answering.}
\add{LLMPhy~\cite{LLMPhy_2025} goes one step further by coupling language models with world-model components, suggesting that symbolic reasoning can be strengthened when the model can internally simulate latent physical dynamics instead of relying only on textual heuristics.}

\add{\noindent\textbf{Evaluation Landscape.} }
\add{Symbolic reasoning is primarily evaluated on text-based benchmarks such as PhysicsEval~\cite{physicseval}, UGPhysics~\cite{ugphysics}, PHYBench~\cite{qiu2025phybench}, ABench-Physics~\cite{abenchphysics}, and GPQA~\cite{GPQA}.} \add{These benchmarks jointly measure final-answer accuracy, multi-step consistency, and robustness on harder, less memorization-driven physics problems.} \add{The results indicate that even the strongest current reasoning models still fall short of human experts (Gemini 2.5 Pro scores 36.9 vs. 61.9 for humans)~\cite{qiu2025phybench}.}

\subsection{\add{Multimodal-grounded Reasoning}}
\add{\noindent\textbf{Methods.}}
On another dimension, a fundamental distinction between physics reasoning and pure mathematics lies in the involvement of more complex visual perception and diagram comprehension, which has driven the development of multimodal methods. 
\add{Multimodal-grounded reasoning tasks build upon physical perception by first extracting scene-grounded objects, relations, and events from diagrams, images, or videos, and then applying symbolic physical laws to perform reasoning and produce answers.}
\add{This setting is more challenging because the model must bind abstract physical principles to scene-grounded entities and their interactions.}

\add{An influential early example is Dynamic Concept Learner~\cite{chen2021grounding}, which grounds physical objects and events from video and performs reasoning through structured dynamic representations.}
\add{TRACE~\cite{Imani2025TRACE} further shows that multimodal reasoning should be analyzed at the step level, since small grounding errors in early stages can easily propagate into physically inconsistent conclusions.}
\add{For diagram-heavy problems, Liang et al. reduce the modality gap by converting visual content into structured textual descriptions before reasoning, which substantially improves performance on visual physics problem solving~\cite{Liang2025MultimodalRF}.}
\add{For olympiad-style settings that require tighter visual grounding, P1-VL~\cite{luo2026p1} integrates dense visual perception and scientific reasoning within a single model, reducing the dependence on manually engineered modality conversion.}

\add{\noindent\textbf{Evaluation Landscape.} }
\add{This task family is commonly evaluated on PhysReason~\cite{physreason}, CLEVRER~\cite{clevrer}, and ComPhy~\cite{ComPhy}, with MMMU and MMMU-Pro~\cite{mmmu,mmmupro} serving only as broader multimodal stress tests.} \add{On SeePhys, Gemini-2.5-Pro drops by 25.7\% on vision-essential questions relative to vision-optional ones~\cite{seephys}.} \add{These results suggest that the main bottleneck is not simple recognition or symbolic reasoning, but the weak coupling between grounded perception and structured physical reasoning.}

\subsection{Causal and Counterfactual Reasoning}
\add{\noindent\textbf{Methods.}}
\add{Beyond multimodal-grounded reasoning, an important next step is to infer the hidden mechanisms behind observed events.}
\add{Rather than only answering questions from perceptual evidence, causal and counterfactual reasoning asks why an event occurs and how the outcome would change under intervention.}
Representative approaches in this direction explicitly model causal structure or intervention effects, instead of relying only on observational correlations. Causal graph modeling~\cite{causalmodeling} and intervention-based learning~\cite{ke2020causalmodels} provide general frameworks for recovering latent dependencies from dynamic observations. Causal Threads~\cite{causalthreads} further explains state changes through structured causal traces, making the reasoning process more interpretable. In more physics-grounded settings, differentiable-physics-based visual reasoning~\cite{ding2021dynamic} combines learned perception with explicit physical constraints to support mechanistic inference and counterfactual analysis. Compared with standard multimodal reasoning models, these methods are more suitable when the goal is to identify what factor actually drives a physical event. \del{Evaluation of these approaches relies on the development of various physical reasoning benchmarks: the Causal3D benchmark}\del{~\cite{causal3D}}\del{ provides structured data with corresponding visual representations; the CLEVRER-Humans dataset}\del{~\cite{clevrerhumans}}\del{ offers video-based reasoning tasks for causal judgment of physical events with human annotations, addressing the need for diverse event types and natural language descriptions; and the recent PhySense benchmark}\del{~\cite{physense}}\del{ emphasizes principle-based reasoning by systematically evaluating whether large language models can apply human-like physical principles.}

\add{\noindent\textbf{Evaluation Landscape.} }
\add{This capability is commonly evaluated by benchmarks that explicitly target latent causality or intervention-based reasoning.}
\add{CAUSAL3D~\cite{causal3D} is representative in this respect: it contains 19 3D-scene datasets and shows that performance declines markedly as causal structures become more complex and prior knowledge is removed~\cite{causal3D}.}
\add{CLEVRER-Humans~\cite{clevrerhumans} provides human-annotated causal judgments over physical events and reveals a large gap between current models and human reasoning: ALOE achieves only 26.9\% per-question accuracy, whereas humans reach 71.4\%~\cite{clevrerhumans}.}
\add{At the symbolic reasoning level, PhySense~\cite{physense} further tests whether models can solve 380 novel physics problems through concise principle-first reasoning rather than long correlation-driven derivations, and shows that current LLMs still fall short of expert-like reasoning efficiency and interpretability~\cite{physense}.}

\subsection{Accelerate Physics Research}

\subsubsection{Hypothesis Generation}
\noindent\textbf{Methods.}
Data-driven approaches aim to discover unknown physical laws directly from data.
A prominent example is Symbolic Regression (SR), which extracts interpretable mathematical expressions without prior knowledge of the governing equations.
The key challenge is navigating the combinatorial search space while enforcing physical plausibility.
Recent work therefore incorporates hard physical constraints (e.g., unit consistency) and domain priors to guide the search toward meaningful equations~\cite{zhang2024_interpretable_turbulence_sr}.

\add{Recent domain-specific systems suggest a broader agentic pattern for hypothesis generation.
For example, the multi-agent framework of Hu et al. automates variable selection, hypothesis formulation, symbolic regression, and mechanistic explanation for physical-law discovery~\cite{hu2024_multiagent_materials_discovery}.}

\noindent\textbf{Evaluation Landscape.}
\add{Hypothesis-generation systems are commonly evaluated by expression recovery, dimensional consistency, sparsity, extrapolation, interpretability, and physical plausibility~\cite{LaCava2021SRBench,deFranca2025SRBenchPP,mazheika2024_ga_sisso,huang2025_domain_aware_sr_priors,zhang2024_interpretable_turbulence_sr,aravanis2025_asp_sr_fluid}.}

\subsubsection{Streamline Experimentation}
\begin{figure}[t]
    \centering
    \includegraphics[width=0.48\textwidth]{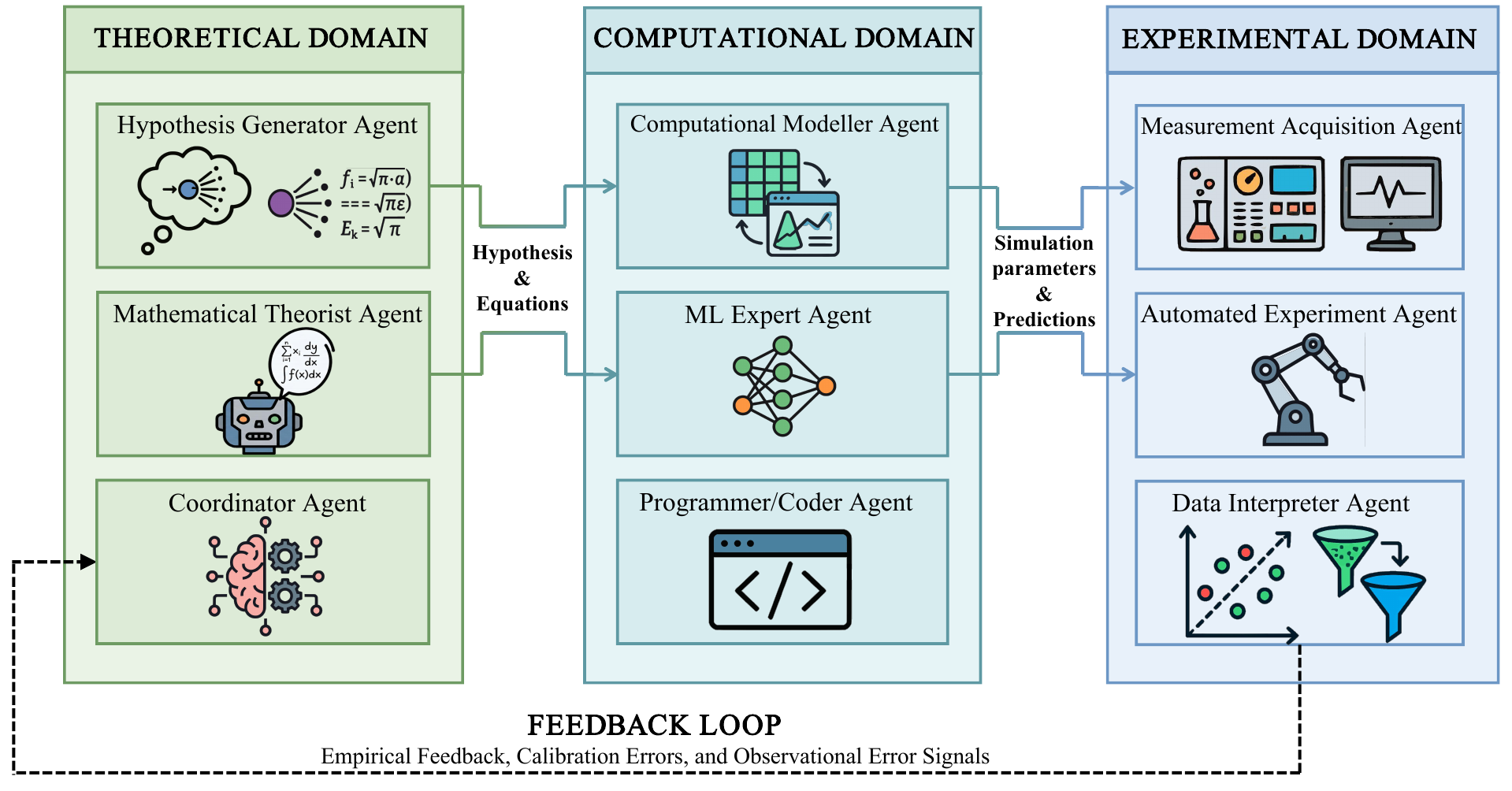} 
    \caption{\del{Multi-agent physics reasoning system across 3 use case spanning theoretical, computational and experimental research.}\add{Multi-agent workflow for physics research automation across theoretical, computational, and experimental domains. Theoretical agents generate hypotheses and equations, computational agents perform modeling and simulation, and experimental agents support measurement, automation, and data interpretation. The arrows indicate the feedback loop from hypothesis generation to prediction and experimental validation.}}
    \label{fig:supernova}
\end{figure}

\noindent\textbf{Methods.}
Beyond generating candidate laws, research acceleration requires tool-using agents that can turn ideas into executable experiments.
The AI Scientist and AI Scientist-v2 provide end-to-end scaffolds for autonomous idea generation, code writing, experiment execution, result analysis, and manuscript drafting~\cite{lu2024aiscientist,yamada2024aiscientistv2}.
In physics-related settings, cmbagent organizes specialized agents for retrieval, coding, critique, and execution, and applies this workflow to cosmology parameter estimation from supernova data~\cite{Xu2025cmbagent}.
Recent domain-specific systems further show how physics reasoning can become a productive scientific tool: AtomAgents supports alloy design via retrieval, multimodal analysis, and physics-based simulation~\cite{ghafarollahi2024atomagents}, while LLM-driven turbulence modeling closes the loop between proposal, simulation-based evaluation, and refinement~\cite{yang2025_llm_turbulence_modeling}.
\add{At the systems level, ScienceClaw+Infinite and ClawdLab point toward persistent, provenance-aware research ecosystems in which agents share artifacts, memory, and verification protocols across longer workflows~\cite{Wang2026ScienceClaw,Weidener2026ClawdLab}. Figure~\ref{fig:supernova} summarizes this multi-agent research automation pipeline.}

\noindent\textbf{Evaluation Landscape.}
Evaluation should not be reduced to answer accuracy alone.
More informative criteria include end-to-end task completion, correctness of generated code and simulations, scientific validity and interpretability of produced artifacts, and whether the system can sustain iterative tool use under long-horizon memory.
ScienceAgentBench highlights the difficulty: even the best evaluated agent solves only 32.4\% of tasks independently and 34.3\% with expert-provided knowledge~\cite{scienceagentbench}.
Complementary benchmarks such as SciCode further probe whether agents can reliably translate scientific intent into executable research code~\cite{scicode}.
\add{These results suggest that physics research automation is now feasible in narrow, tool-rich settings, but robust open-ended scientific autonomy remains far from solved.}

%% file: sec/5_world_models.tex
\section{World Modeling}
\label{sec5}


The previous sections have traced two key paths, namely Physical Perception and Physics Reasoning, towards AI systems that perceive and understand the law of physics.
Physical perception embodies the end-to-end learning approach, extracting physical patterns and regularities directly from sensory observations, while physics reasoning encodes scientific knowledge into symbolic representations and theoretical frameworks.

After internalizing such physical knowledge, can models strengthen their understanding of governing principles of the world and thereby predict the future? World models, as an emerging line of research, offer a natural next step for physical AI systems by enabling a progression from understanding to modeling.

\textbf{World Modeling} represents the capability of AI systems to generate, understand and predict the dynamics of simulated environments, enabling a wide range of applications from video generation and scene reconstruction to autonomous planning and decision-making.

Models with world modeling ability have three fundamental advantages for current AI development: (1) reducing the need for supervised data, (2) closing the simulation-to-reality gap, and (3) enabling low-cost, interpretable prediction of future states.
These advantages arise because world models enable agents to learn through interaction with simulated environments rather than relying on massive amounts of data, and they can provide more fine-grained sensory data for embodied agents~\cite{understanding_or_pridiction}.

Unlike physics reasoning\del{ LLMs, world models exhibit substantial shifts in their architectural design, training paradigms, and evaluation methodologies. In this section, we categorize litertures of world models into three dimensions: \textit{(1) Generative Models:} They rely on gradient descent algorithm to learn underlying dynamics of the world from large scale database; \textit{(2) Physics-enhanced Approaches:} They are constructed based on the principles and laws of physics in symbolic representation, combining the principle of physics that govern our physical world with neural networks to simulate changes in world states; \textit{(3) Benchmarks:} They provide the metrics used to evaluate model's capabilities in qualitative and quantitative estimation for world modeling.}\add{, world modeling is defined by predicting, reconstructing, or simulating how physical environments evolve. Specifically, we consider four task families: image generation, video generation, scene reconstruction, and physics-constrained simulation. For each task, we first review representative methods and then summarize the corresponding evaluation landscape.}

\subsection{\del{Generative Models}\add{Image Generation}}

\del{\noindent\textbf{Image Generation. }}
\add{\noindent\textbf{Methods.}}
Image generation models transform abstract information into static visual representations.
This process encompasses not only traditional lighting models like Phong Shading~\cite{Phong} and material rendering like Cook-Torrance~\cite{Cook-Torrance}, but more crucially utilizes \textit{Physically Based Rendering (PBR)} through physical simulators like NVIDIA Isaac Sim~\cite{isaacsim} and Unity ML-Agents~\cite{Unity-ML-Agents} to simulate the interactions between light rays and objects, thereby producing images with high fidelity in both visual and physical perspectives.
Recent work in computer vision has demonstrated the effectiveness of PBR towards closing the sim-to-real gap, with applications ranging from depth sensor simulation ~\cite{stereovision-depth-sensors}\del{ to 6D object pose estimation ~\cite{BOP}, where PBR-generated training data significantly outperformed traditional rendering approaches}\add{ to sim-to-real visual transfer for downstream perception and control}.

Under the world model framework, image synthesis establish the connection between internal physical state representations and external visual observations. Contemporary neural rendering approaches, especially those based on Neural Radiance Fields (NeRF), such as Ref-NeRF~\cite{Ref-NeRF} and ENVIDR~\cite{ENVIDR}, allow the rendering process to produce both visually convincing images and physically consistent results.
This transformation process from physical simulation to image synthesis effectively translates the internal state representations of world models into an observable and interpretable form for embodied agents, establishing a critical visual foundation for temporal modeling and dynamic prediction.

\add{\noindent\textbf{Evaluation Landscape.} }
\add{Compared with later world-modeling tasks, evaluation for image generation remains dominated by fidelity and consistency metrics such as FID~\cite{FID} and CLIP Score~\cite{Clipscore}\del{, typically measured on datasets such as COCO-30K~\cite{coco} and ImageNet~\cite{Imagenet}}.}
\add{When physical realism is important, additional tests examine whether generated observations preserve geometry or lighting consistency, for example on NYU Depth V2~\cite{NYU-Depth-V2}.}
\add{This suggests that image generation is a useful grounding task for world models, but its evaluation remains less physics-specific than that of video generation or simulation.}

\subsection{\add{Video Generation}}
\del{\noindent\textbf{Video Generation. }}
\add{\noindent\textbf{Methods.}}
Video generation refers to the computational technique of automatically creating continuous video sequences from input data such as text, images, or noise using algorithmic methods. As we extend from static image synthesis to dynamic video generation, physical constraints become increasingly complex and important.

\begin{figure}[t]
    \centering
    \includegraphics[width=0.48\textwidth]{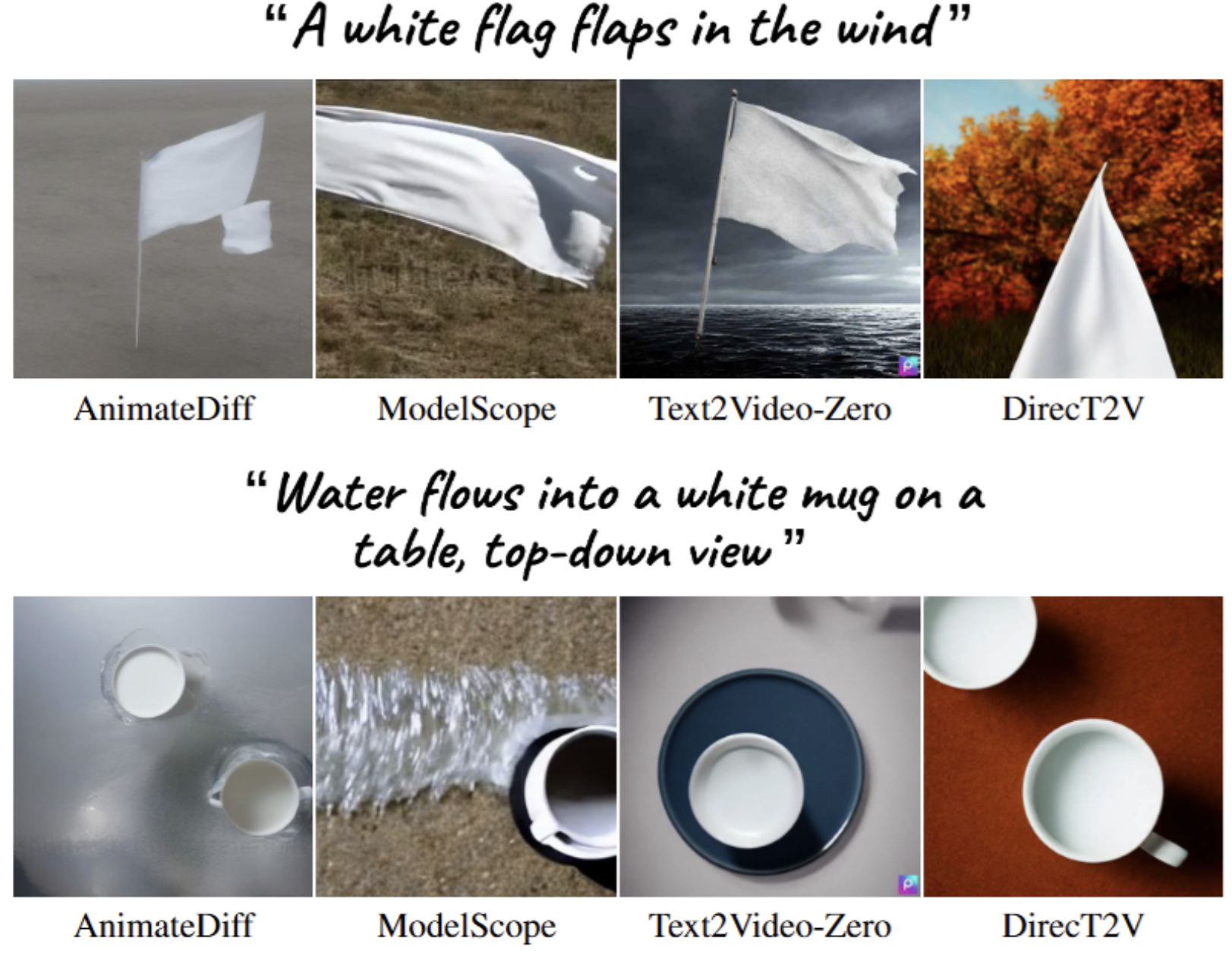} 
    \caption{Video Generations by GPT4Motion (Figure courtesy of~\cite{GPT4Motion}).
    }
    \label{fig: GPT4Motion}
\end{figure}

\begin{table}[t]
\centering
\caption{Performance of Video Generation Models on representative physical and world modeling benchmarks.}
\label{tab:vgen_bmk}

\begin{tabularx}{\linewidth}{>{\raggedright\arraybackslash\hsize=0.9\hsize}X
                             >{\centering\arraybackslash\hsize=0.7\hsize}X
                             >{\centering\arraybackslash\hsize=0.6\hsize}X
                             >{\centering\arraybackslash\hsize=0.6\hsize}X
                             >{\centering\arraybackslash\hsize=0.9\hsize}X}
\toprule
\textbf{Model} & \textbf{\shortstack{PhysicsIQ\\\cite{Physics-IQ}$(\uparrow)$}} & \textbf{\shortstack{PhyGen\\\cite{phygen}$(\uparrow)$}} & \textbf{\shortstack{VideoPhy\\\cite{2024videophy} $(\uparrow)$}} & \textbf{\shortstack{WorldModel\\Bench\cite{worldmodelbench}$(\uparrow)$}} \\
\midrule
Sora~\cite{opensora}           & 0.10 & 0.44 & 0.28 & 6.11 \\
Pika      & 0.13 & 0.44 & 0.29 & --  \\
CogVideoX~\cite{CogVideoX}   & -- & 0.45 & 0.49 & 7.31  \\
LaVie~\cite{wang2023lavie}       & -- & 0.36 & 0.41 & --  \\
Kling       & -- & 0.49 & -- & 8.82  \\
\bottomrule
\end{tabularx}
\end{table}

Commercial product webpages in Table~\ref{tab:vgen_bmk} are cited as footnotes rather than bibliography entries: \webref[Pika official page]{Pika}{https://pika-art.net/}{Apr. 13, 2026} and \webref[Kling official page]{Kling}{https://kling.ai/app/}{Apr. 13, 2026}.

Modern video generation models, such as diffusion-based approaches like GPT4Motion~\cite{GPT4Motion} and VLIPP~\cite{VLIPP}\add{, together with physics-guided variants such as PhysGen~\cite{PhysGen}, ProPhy~\cite{Wang2025ProPhyPP} and DiffPhy~\cite{DiffPhy},} are progressively constrained by physics laws, making the generated video sequences not only visually coherent but also physically plausible. For example, as illustrated in Figure~\ref{fig: GPT4Motion}, GPT4Motion is constrained by physics laws through physical simulators like \webref[PyBullet project page]{PyBullet}{https://pybullet.org/wordpress/}{Apr. 7, 2026} by using GPT-4 to generate Blender scripts that simulate realistic physics dynamics, ensuring both visual consistency and adherence to physical laws across video frames, while VLIPP employs chain-of-thought reasoning in vision-language models to predict physically plausible motion trajectories that guide diffusion models to generate temporally coherent and physically accurate video sequences. DINO-Foresight~\cite{Karypidis2024DINOForesightLI} supports multiple future understanding tasks and avoids reconstructing irrelevant details of pixel based methods.

This physics-constrained video generation capability is crucial for building reliable world models, as it serves as \del{an evaluation metric of world models' understanding of environmental dynamics. As shown in Table~\ref{tab:vgen_bmk}, recent video generation models demonstrate varying performance across physical and world modeling benchmarks, indicating their capability to capture environmental dynamics. The ability to generate physically plausible video sequences indicates that the underlying world model has developed robust representations of environmental dynamics, making such generation capability a valuable benchmark for assessing world model quality.}\add{an observable proxy for whether the model has captured environmental dynamics.}

\add{\noindent\textbf{Evaluation Landscape.} }
\add{As shown in Table~\ref{tab:vgen_bmk}, current video generation models still exhibit a clear gap between visual fluency and physical consistency.}
\add{Kling achieves the best reported results on PhyGen (0.49)~\cite{phygen} and WorldModelBench (8.82)~\cite{worldmodelbench}, whereas CogVideoX~\cite{CogVideoX} performs best on VideoPhy (0.49)~\cite{2024videophy}.}
\add{Performance on PhysicsIQ remains particularly low: the best reported score in the table is 29.5, achieved by VideoPoet~\cite{VideoPoet} on PhysicsIQ~\cite{Physics-IQ}, suggesting that learning robust physical principles from video is still substantially harder than producing visually coherent motion.}
\add{Morpheus~\cite{Morpheus} complements these synthetic benchmarks with real physical experiments, highlighting the remaining gap between benchmark optimization and real-world physical fidelity.}
\add{Overall, these results indicate that current video world models capture benchmark-specific aspects of dynamics, but still lack uniformly strong physical consistency across evaluation settings.}

\subsection{\add{Scene Reconstruction}}
\del{\noindent\textbf{Scene Reconstruction. }}

\add{\noindent\textbf{Methods.}}
Scene reconstruction is a computer vision technique that recovers the three-dimensional geometric structure and appearance information of a scene from two-dimensional images, point clouds, or other sensor data, aiming to construct a complete digital 3D representation of the scene. Unlike the forward process discussed earlier that transforms physical states into visual representations, scene reconstruction is an inverse process that requires inferring complete 3D geometry, physical properties, and scene dynamics from limited observational information.

Traditional 3D reconstruction methods are typically based on geometric methods like SFM~\cite{SFM} and MVS~\cite{MVS}, so they demonstrate strong geometric perception capabilities and often neglect the significance of physical reasoning, resulting in reconstructed scenes that violate fundamental physical principles.

Under the world model framework, 3D scene reconstruction should not only recover static geometric structures, but also understand and reconstruct the physical properties and dynamic behaviors among different objects within the scene. This requires reconstruction algorithms to possess not only geometric perception capabilities, but also physical reasoning capabilities.
Modern scene reconstruction methods, particularly neural implicit representation-based approaches like PhyRecon~\cite{PhyRecon} and IDR~\cite{IDR}\add{, together with world-model-driven systems such as ReconDreamer~\cite{ReconDreamer} and DriveDreamer4D~\cite{DriveDreamer4D},} are making progress in this direction. Not only can they reconstruct precise geometric structures, but also learn and encode physical properties such as density, stiffness, and friction coefficients through physical simulators like MuJoCo~\cite{MuJoCo} and \webref[Warp project page]{Warp}{https://developer.nvidia.com/warp-python}{Apr. 7, 2026}. This capability enables reconstructed scenes to be not only visually realistic but also accurate in physical behavior, which is precisely the key requirement for building actionable world models.

\add{\noindent\textbf{Evaluation Landscape.} }
\add{Evaluation in this task family typically combines geometric accuracy, completeness, and novel-view synthesis quality.}
\add{Classical benchmarks such as the DTU multi-view stereo benchmark~\cite{DTU2014}, which contains 80 scenes with reference structured-light scans, and Tanks and Temples~\cite{TanksAndTemples2017}, which evaluates large-scale indoor and outdoor reconstruction against laser-scanner ground truth, remain standard testbeds for geometry fidelity under realistic viewpoint variation.}
\add{ETH3D~\cite{ETH3D2017} further extends evaluation to higher-resolution and mobile-view settings.}
\add{For driving-oriented 4D reconstruction, DriveDreamer4D~\cite{DriveDreamer4D} further evaluates novel-trajectory rendering with FID~\cite{FID} and reports NTA-IoU for spatiotemporal agent coherence.}
\add{Overall, current evaluation is still dominated by geometric reconstruction quality, while physics-aware reconstruction remains much less standardized.}

\subsection{\del{Physics-enhanced Modeling Approaches. }\add{Physics-constrained Simulation}}
\add{\noindent\textbf{Methods.}}
\add{Physics-constrained simulation treats the output of the world model as future trajectories, object interactions, or controllable rollouts that must remain compatible with mechanics, contact, and conservation laws.}

The core challenge facing existing world models is \textbf{insufficient out-of-distribution physical reasoning capability}. While large-scale systems like \del{GAIA-1~\cite{GAIA-1} and }DINO-world~\cite{DINO-world2025}\del{ and DriveDreamer4D~\cite{DriveDreamer4D}} excel at in-distribution tasks and can generate realistic future scenarios, when encountering counterfactual reasoning, novel object interactions, or unfamiliar physical constraints, these models rely on "case-based" rather than "rule-based" generalization, leading to dramatic performance degradation~\cite{physical-law-perspective2024}. Although early physics-based simulation methods can maintain consistency with real-world physics through predefined physics engines\del{~\cite{differentiable-stokes-flow2020}~\cite{Scalable-differentiable-physic2020}}\add{~\cite{Brax,NeoPhysIx}}, their reliance on manually designed parameterization limits their generalization capability to unmodeled physical effects.

To achieve genuine physical understanding, world models need to accomplish the following critical tasks: (1) Counterfactual physical reasoning: accurately predicting physical behaviors in unseen scenarios; (2) Novel interaction modeling: handling complex inter-object interactions that did not appear during training; (3) Symbolic-grounded knowledge bridging: connecting abstract physical laws with concrete perceptual experiences; (4) Physical constraint generalization: maintaining reasoning accuracy under new constraint conditions. These tasks require systems to maintain strict adherence to fundamental physical principles while adapting to complex and dynamic real-world environments.

Physics-constrained world modeling methodologies merge explicit physical computations with neural network learning that captures complex material behaviors through data-driven optimization, while preserving end-to-end differentiability. Current approaches can be categorized into three main paradigms: 

\add{Firstly,} neuro-symbolic integration directly embeds known physical laws into the neural network architecture to ensure strict compliance with fundamental physical principles, with the neural network solely responsible for learning uncertain components that are challenging to model accurately. \del{For example, PhysORD~\cite{PhysORD2024} models vehicles as controlled Lagrangian systems by directly embedding the Euler-Lagrange equations as hard constraints within the network, where the neural network only needs to estimate quantities that are difficult to compute accurately, such as potential energy gradients and external forces. This approach ensures strict enforcement of fundamental physical laws like energy conservation, thereby maintaining good generalization capability even under data-scarce conditions.}\add{For example, SAIN~\cite{SAIN2019} combines object-centric neural predictions with an explicit physical simulator, so that learned representations can be corrected by structured dynamics during control. Related neuro-symbolic efforts such as DEM-NeRF~\cite{tan2025_dem_nerf_neurosymbolic} further show how explicit physical constraints can regularize neural scene evolution.}

\add{Furthermore,} physics-structured neural oDEs retain the mathematical structure of physical equations in the neural ODE framework, where neural networks parameterize various components of the physical equations to enable continuous-time domain modeling. For example, MoSim~\cite{MoSim2025} decomposes the rigid body dynamics equations into predictor and corrector parts. The predictor strictly follows the mathematical structure of rigid body dynamics, with inertia matrices, gravity terms, and control forces parameterized through specialized neural network modules. The corrector uses standard residual networks to address complex phenomena such as friction and collisions that are challenging to model explicitly. Time integration is performed via neural ODE solvers, enabling the system to maintain both the inherent structure of physical equations and the learning capacity of neural networks. \add{LagNetViP~\cite{lagnetvip} similarly preserves Lagrangian structure for video prediction, showing that continuous-time inductive bias can improve rollout stability.}

\add{Finally,} differentiable physics engines employ a decoupled architecture in which neural networks specialize in predicting physical properties and interaction parameters from sensor data, with these predictions subsequently input into differentiable classical physics solvers for final system state computation. FusionForce~\cite{fusionforce2025} utilizes deep neural networks to predict terrain physical properties such as geometric shape, friction coefficients, and stiffness, along with robot-terrain contact forces from camera images and LiDAR data. These predictions are subsequently fed into a differentiable rigid body dynamics solver for robot motion trajectory computation. The end-to-end differentiable framework allows the model to refine terrain property predictions by backpropagating trajectory errors.

These methodologies demonstrate how physical reasoning emerges from the synergistic interaction between neural learning and constraints offered by the law of physics, moving beyond pattern memorization toward principled understanding of environmental dynamics.

\add{\noindent\textbf{Evaluation Landscape.} }
\add{Physics-constrained simulators are judged not only by perceptual fidelity but also by whether their rollouts remain useful under intervention, planning, and distribution shift.}
\add{Benchmarks such as PHYRE~\cite{phyre} test intervention-based physical generalization, while PhyWorldBench~\cite{PhyWorldBench} evaluate whether simulated rollouts remain physically realistic under diverse scenarios.}
\add{In practice, however, evaluation is still fragmented: control-oriented systems such as SAIN~\cite{SAIN2019} and FusionForce~\cite{fusionforce2025} report downstream control success or trajectory error, while RL world models such as MoSim~\cite{MoSim2025} emphasize long-horizon return and rollout accuracy.}
\add{Taken together, these benchmarks and task-specific protocols emphasize state-transition accuracy, intervention robustness, and long-horizon physical consistency rather than visual quality alone.}

%% file: sec/6_embody_ai.tex
\section{Embodied Interaction}
\label{sec6}


World models provide the possibility for modeling virtual physical scenarios, yet a substantial gap persists between simulation environments and the real world. This gap highlights an essential limitation: modeling alone is insufficient for mastering the dynamics of real physical processes. Embodied intelligence addresses this limitation by requiring agents to perform multi-dimensional reasoning and interact with their physical surroundings. As a result, it moves beyond approaches that operate solely within controlled or virtual spaces and is increasingly viewed as a key paradigm for advancing future physics-based AI.


\subsection{Robotics}
\add{\noindent\textbf{Methods.}}
While conventional world models excel at predicting outcomes in simulated environments, robotics applications demand the critical transition from passive observation to active manipulation. This evolution is exemplified by the emergence of Vision-Language-Action (VLA) models~\cite{VLA} that bridge the gap between building internal world representations and changing external physical reality.
Pioneering systems like Gato~\cite{ReedEtAl2022_Gato} and RT-1/2~\cite{BrohanEtAl2022_RT1,Brohan2023RT2VM} establish the foundation for this paradigm shift.

\begin{figure}[t]
    \centering
    \includegraphics[width=0.48\textwidth]{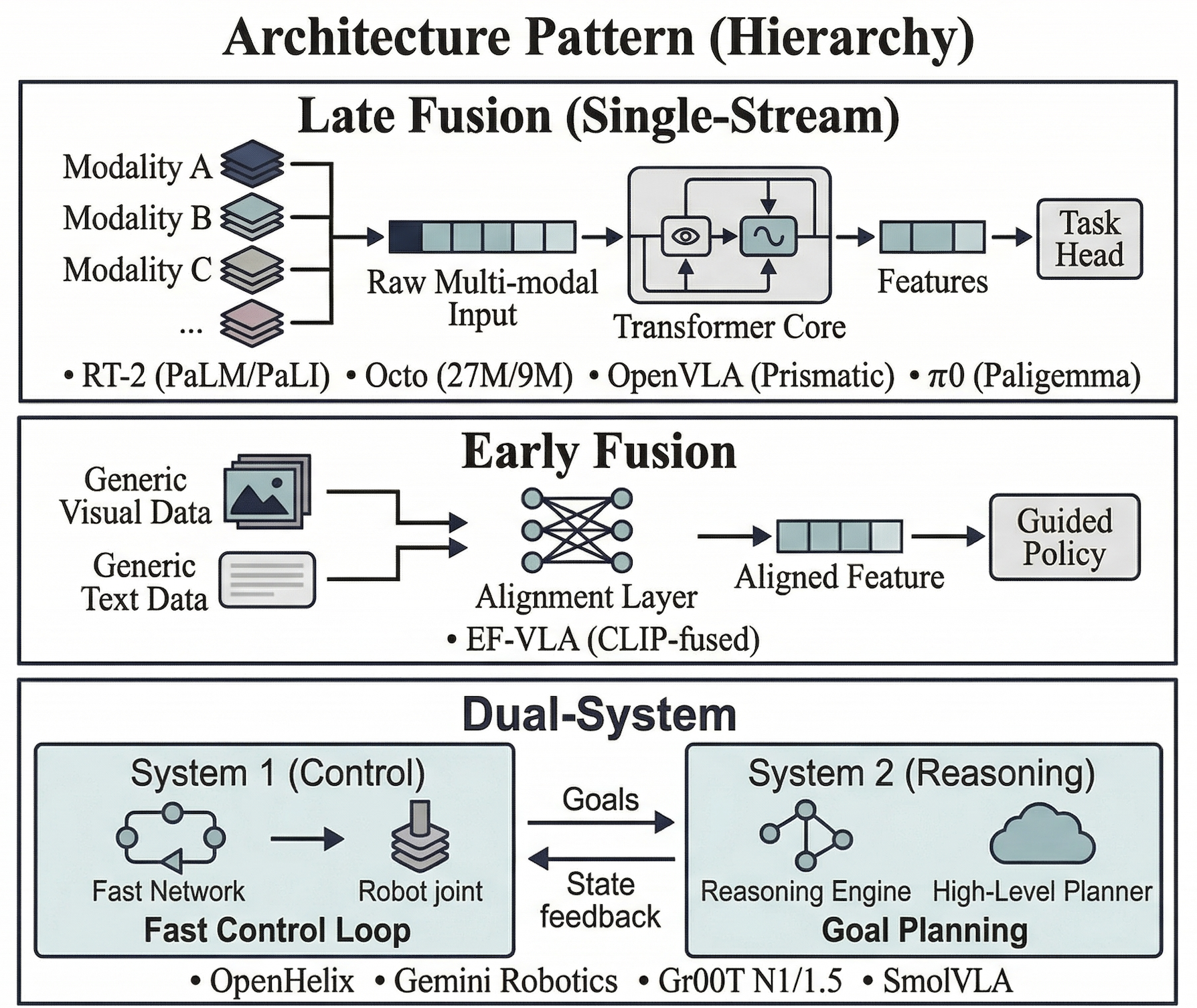} 
    \caption{\del{Architecture patterns for vision-language-action models. The shift from Late Fusion single-stream architectures to Dual-System designs reflects the integration of physical reasoning into VLAs.}\add{Representative architecture patterns for vision-language-action models. Top: late-fusion single-stream models fuse multi-modal inputs in a shared transformer before action prediction. Middle: early-fusion models align vision and language before policy generation. Bottom: dual-system models separate fast control from deliberative reasoning and planning. This comparison highlights the difference between fusion-based and modular reasoning-augmented VLA designs.}
    }
    \label{fig:architecture_pattern}
\end{figure}

Recent advances, including $\pi_0$~\cite{BlackEtAl2024_pi0}, OpenVLA~\cite{KimEtAl2024_OpenVLA}, and \webref[Gemini Robotics official blog post]{Gemini Robotics}{https://deepmind.google/discover/blog/gemini-robotics-brings-ai-into-the-physical-world/}{Apr. 13, 2026} further demonstrate how physics-enhanced AI with embodiment extends beyond simulation-based world models by requiring real-time adaptation to physical constraints, sensor noise, and the irreversible consequences of actions in the real world. Figure~\ref{fig:architecture_pattern} illustrates primary paradigms for current VLA systems. Both fusion-based (late/early) and dual-system architectures reflect the importance of integrating fast physical control with deliberative reasoning.

The transition from simulated world models to reality systems also introduces new challenges. First, continuous action generation highlights the inadequacy of discretized representations when confronting the smooth, continuous nature of reality, because robotics demands continuous control that respects physical constraints~\cite{Safety-Critical}. Models like $\pi_0$ address this through flow matching~\cite{lipman2023_flow_matching}, generating physically plausible trajectories that maintain stability during contact-rich manipulation.
This direction is further explored through diffusion policies~\cite{ChiEtAl2023_DiffusionPolicy} . Furthermore, cross platform generalization tests whether learned physical principles can transcend specific hardware implementations. OpenVLA's approach of learning normalized action deltas from diverse datasets~\cite{BrohanEtAl2022_OpenX} demonstrates how embodied model must abstract beyond simulation specific assumptions to achieve morphology independent reasoning that generalizes across varied physical platforms. \add{Recent work on World Action Models (WAMs) further demonstrates that predicting future world states facilitates a more robust understanding of physical dynamics, thereby enhancing closed-loop robotic control in complex real-world scenarios~\cite{Ye2026WorldAM,Liu2026WorldAV,Liu2026DriveVAVA}.} Third, real-world perceptual grounding requires models to construct accurate world representations from noisy, partial observations. Some works addresses this through enhanced multi-view correspondence understanding in Gemini Robotics, object localization~\cite{LingEtAl2023_ActiveLearning,SeoEtAl2023_HybridControl} and physical properties recognition~\cite{WuEtAl2023_ObjectProperty}, capabilities essential for reasoning about physical interactions under perceptual uncertainty. 

\add{\noindent\textbf{Evaluation Landscape.} }
\add{Many robotics tasks offers diverse established evaluation suites spanning physical perception, manipulation and navigation for embodied AI. EmbodiedBench~\cite{EmbodiedBench} offers a broad diagnostic across planning, navigation, and manipulation for vision-driven embodied agents, while EMMOE~\cite{EMMOE} emphasizes long-horizon mobile manipulation in open environments. For manipulation, RLBench~\cite{RLBench2020}, RoboSuite~\cite{RoboSuite2020}, ManiSkill2~\cite{ManiSkill2}, Meta-World~\cite{MetaWorld2019}, CALVIN~\cite{CALVIN2022}, and BEHAVIOR~\cite{BEHAVIOR2021} provide standardized tasks with diverse objects, goals, and contact conditions, enabling repeatable evaluation and meaningful comparison across methods. Cross-platform robustness is often stress-tested via heterogeneous training and transfer on Open X-Embodiment~\cite{BrohanEtAl2022_OpenX}, which checks whether action representations remain grounded across different robots, tasks, and sensing conditions. On the perception side, benchmarks such as BOP for 6D object pose estimation~\cite{BOP} and GraspNet-1Billion for grasp pose prediction in RGB-D scenes~\cite{GraspNet2020} assess whether models can recover geometry, pose, and affordances from real sensors.}

\subsection{Navigation}

\del{Navigation, a cornerstone of robotics, has evolved from a classical geometric problem to a core challenge in embodied intelligence, demanding interaction with and reasoning about the physical world. Current representative navigation tasks include object-goal navigation, vision-and-language navigation, dialog-based navigation, etc.}

\del{\noindent\textbf{Object-Goal Navigation. }
Object-Goal Navigation (ObjectNav) tasks an agent to interact with its physical world by exploring an unknown space to find a target object, requiring it to reason about potential object locations based on visual cues. It's best represented by the Habitat-Matterport3D (MP3D)~\cite{Matterport3D} benchmark, and benchmarks like AI2-THOR~\cite{AI2-THOR}, Gibson~\cite{Gibson}, RoboTHOR~\cite{RoboTHOR}, HM3D-OVON~\cite{HM3D-OVON} belong to this task. }

\del{\noindent\textbf{Vision-and-Language Navigation.} Vision-and-Language Navigation (VLN) challenges an agent to reason about the connection between language and the physical world, interacting with its environment by translating natural language commands into a specific navigation path. For example, R2R~\cite{mattersim,Matterport3D}, RxR~\cite{rxr}, REVERIE~\cite{REVERIE} and TOUCHDOWN~\cite{Touchdown} are benchmarks for this task. }

\del{\noindent\textbf{Dialog-based Navigation. }
Dialog-based Navigation allows an agent to interact with a user and its physical world simultaneously, reasoning about ambiguous instructions through conversation to successfully navigate its environment. Related to it are RobotSlang~\cite{RobotSlang}, R2H~\cite{R2H}, CVDN~\cite{CVDN}, UNMuTe~\cite{UNMuTe} and so on.}

\noindent\textbf{\del{Evolution of Navigation }Methods. }
Navigation tasks have seen a significant methodological evolution, moving from early, specialized systems to models powered by large language models (LLMs). Early non-LLM-based methods were typically trained on domain-specific data to create a direct mapping from sensory inputs to navigation actions~\cite{zhang2024visionandlanguagenavigationtodaytomorrow}. This reliance on implicit representations often led to models that learned spurious correlations between inputs and outputs, limiting their generalization to unseen environments and making their decisions difficult to interpret~\cite{zhang2025researchnavigationmethodsbased}. 

\begin{figure}[t]
    \centering
    \includegraphics[width=0.48\textwidth]{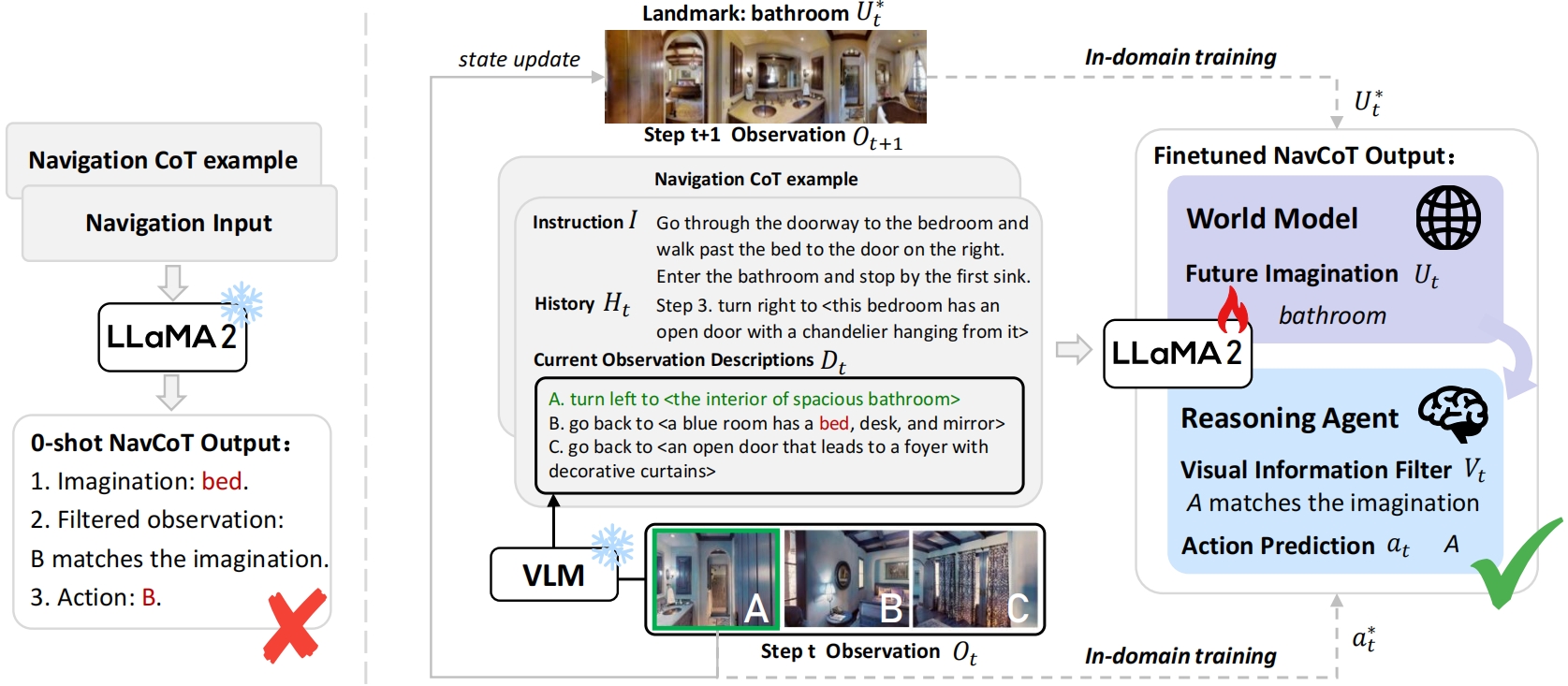} 
    \caption{Overview of NavCoT (Figure used courtesy of~\cite{NavCoT}).
    }
    \label{fig:navcot}
\end{figure}

\noindent The advent of LLMs and VLMs has introduced a new paradigm. Pre-trained on vast, diverse datasets, these models bring a wealth of real-world commonsense and physical knowledge that dramatically improves navigation precision. This shift has enabled explicit reasoning, such as CoT~\cite{CoT}, which not only enhances decision accuracy but also provides greater interpretability~\cite{zhang2025researchnavigationmethodsbased}. This has led to a new wave of methods which are shown in Table~\ref{tab:nav_comparison} and Figure~\ref{fig:navcot}.

\begin{table}[t!]
\centering
\caption{Representative methods for navigation tasks: classical vs. recent multimodal approaches}
\label{tab:nav_comparison}
\begin{tabularx}{\columnwidth}{c|c c}
\toprule
\diagbox{Task}{Method} & \textbf{Classical} & \textbf{Recent multimodal} \\
\midrule
ObjectNav & \makecell{ GOSE~\cite{chaplot2020objectgoalnavigationusing}\\RIM~\cite{RIM}} & \makecell{LOAT~\cite{LOTA}\\ LGR~\cite{LGR}\\ CL-CoTNav~\cite{CL-CoTNav}  \\ASCENT~\cite{ASCENT}} \\
\midrule
VLN & \makecell{BabyWalk~\cite{BabyWalk} \\  ADAPT~\cite{ADAPT} \\ HAMT~\cite{HAMT} \\ EnvDrop~\cite{EnvDrop} \\ DUET~\cite{DUET}} & \makecell{NavGPT~\cite{NavGPT} \\ NavCoT~\cite{NavCoT} \\
VELMA~\cite{VELMA}\\  NaVILA~\cite{NaVILA} }\\
\midrule
Dialog-based & \makecell{CMN~\cite{CMN} \\ GVDN~\cite{GVDN}} & \makecell{\del{ContextBridge} \\ FLAME~\cite{FLAME} \\ \add{UNMuTe~\cite{UNMuTe}}} \\
\bottomrule
\end{tabularx}
\end{table}

\add{\noindent However, these LLM-based approaches are not without their issues.}
\add{They are susceptible to hallucinations, where incorrect physical reasoning can lead to failures, such as imagining a non-existent path through an obstacle~\cite{deng2025llmgoodpathplanner}.}
\add{This fragility is further demonstrated by research showing that minor changes to a prompt can reduce a robot's task success rate by nearly 20\%~\cite{wu2025uncoveringfragilitytrustworthyllms}.}
\add{A fundamental limitation is the representational bottleneck caused by converting rich physical world data into a token format, which can result in a significant loss of detail and the inability to reason about low-level preconditions for actions~\cite{RIM,chaplot2020objectgoalnavigationusing}.}
\add{Recent navigation systems therefore increasingly incorporate predictive world-model components or latent dynamics modules, so that the agent can mentally simulate action outcomes before execution; X-MOBILITY~\cite{X-MOBILITY} is a representative example of this trend.}
\add{The same idea is also essential for bridging the "sim-to-real" gap, as shown by the TWIST~\cite{TWIST} framework's teacher-student world-model distillation strategy.}

\add{\noindent\textbf{Evaluation Landscape.} }
\add{Navigation is evaluated across complementary settings.}
\add{ObjectNav benchmarks such as AI2-THOR, RoboTHOR, Gibson, and HM3D-OVON measure goal-reaching success in unseen environments~\cite{AI2-THOR,RoboTHOR,Gibson,HM3D-OVON}.}
\add{VLN benchmarks including R2R, RxR, VLN-CE, REVERIE, and Touchdown test instruction following under both graph-based and continuous-control regimes~\cite{mattersim,rxr,VLN-CE,REVERIE,Touchdown}, while dialog-based benchmarks such as RobotSlang, CVDN, and UNMuTe probe whether agents can resolve ambiguity through interaction~\cite{RobotSlang,CVDN,UNMuTe}.}
\add{More recent suites such as EmbodiedBench and NavBench further stress long-horizon reasoning, robustness, and cross-domain transfer~\cite{EmbodiedBench,NavBench}.}

\subsection{Autonomous Driving}
\add{\noindent\textbf{Methods.}}
Autonomous driving refers to a system's capability to perform part or all of the dynamic driving task (DDT) on a sustained basis without direct human intervention. The DDT includes all of the real-time operational and tactical functions required to operate a vehicle, such as steering, acceleration, and braking. To achieve this, an autonomous vehicle's closed-loop software system integrates a series of core functions: perception, prediction, planning, and control~\cite{thorn2018framework}. Perception involves collecting and processing real-time data from various sensors like cameras and LiDAR to detect objects and road conditions~\cite{Grigorescu_2019}. Prediction forecasts the behavior of other road users~\cite{Rudenko_2020}. Planning generates a safe and efficient path for the vehicle to follow. Control translates the planned path into physical commands for the vehicle's actuators~\cite{paden2016surveymotionplanningcontrol}. These tasks are implemented through several architectural paradigms~\cite{bojarski2016endendlearningselfdriving}. Current autonomous driving technology methods mainly include: Rule-based Approaches, Learning-based Approaches, World Models and Generative Approaches and Hybrid Approaches.

\del{\noindent\textbf{Rule-based Approaches. }}
Rule-based approaches rely on deterministic algorithms and explicit domain knowledge, encoding physics and traffic rules in interpretable models. Classical methods include the Intelligent Driver Model (IDM)~\cite{IDM} for car-following, RRT/RRT*~\cite{lavalle1998rapidly} for sampling-based trajectory generation, Model Predictive Control (MPC)~\cite{MPC} for optimization-based planning, and potential-field navigation~\cite{khatib1986real}. They are transparent and safe but struggle in dense and uncertain traffic. \del{Benchmarks include CommonRoad~\cite{CommonRoad}, nuPlan~\cite{nuplan}, nuScenes~\cite{nuscenes}, Waymo Open Dataset (WOD)~\cite{Waymo}, CARLA~\cite{CARLA}, and AirSim~\cite{AirSim}.}

\del{\noindent\textbf{Learning-based Approaches. }}
Learning-based approaches replace rules with neural policies trained from data. Imitation Learning (IL) follows expert trajectories (e.g., ChauffeurNet~\cite{ChauffeurNet}), Inverse Reinforcement Learning (IRL) recovers reward functions from demonstrations~\cite{Maximum}, and Reinforcement Learning (RL) optimizes policies in simulators~\cite{kiran2021deepreinforcementlearningautonomous}. AlphaDrive~\cite{AlphaDrive} and AutoDrive-R2~\cite{AutoDrive} extend RL-based training with physics-informed rewards, reinforcing the role of simulation and safety-oriented benchmarks in this category. More recent methods use transformer-based forecasting (e.g., LaneGCN~\cite{LaneGCN}, Wayformer~\cite{Wayformer}).

\begin{figure}[t]
    \centering
    \includegraphics[width=0.48\textwidth]{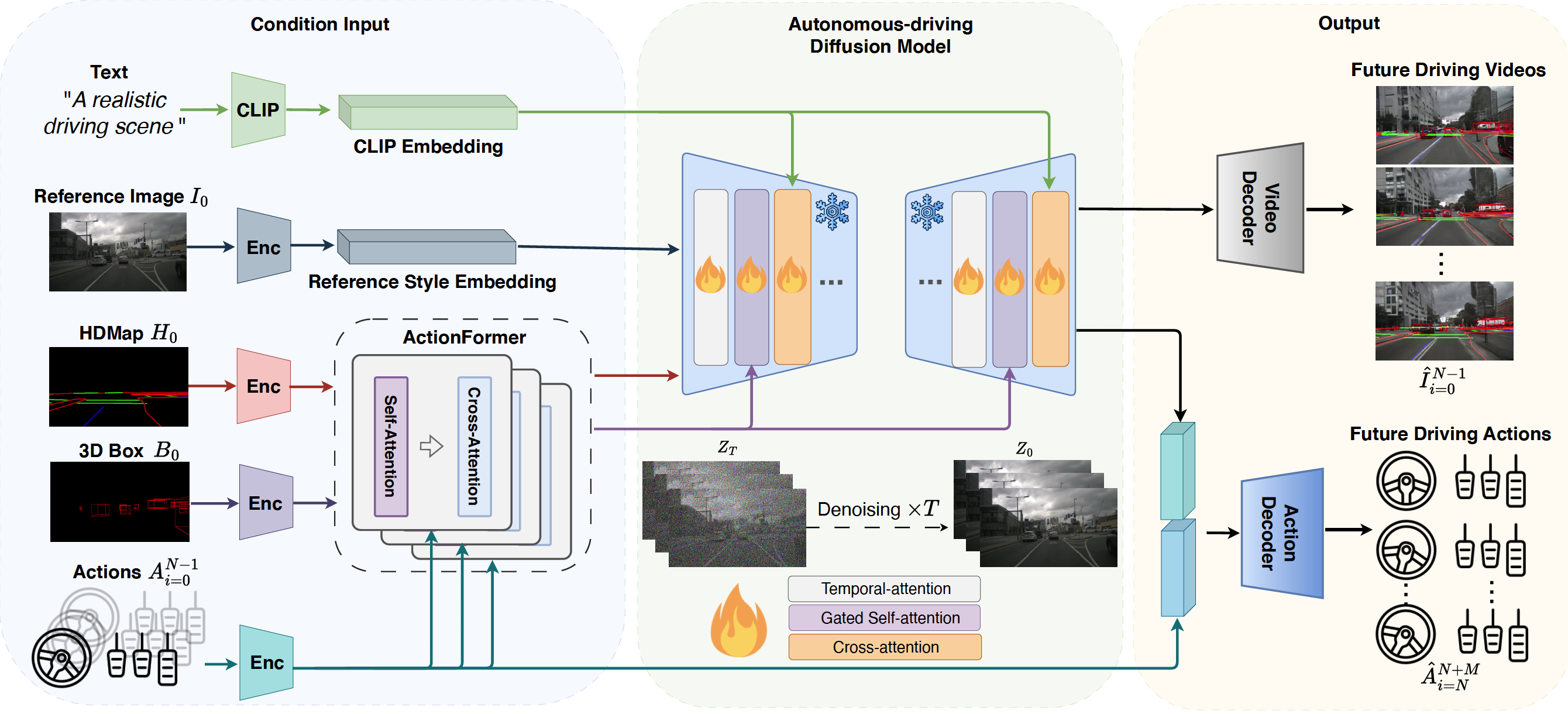} 
    \caption{\del{Overall framework of DriveDreamer}\add{An example driving world model that can serve as an auxiliary simulator for autonomous driving} (Figure used courtesy of~\cite{DriveDreamer}).
    }
    \label{fig:drive_dreamer}
\end{figure}

\del{\noindent\textbf{Generative Approaches. }}
World models learn latent spatio-temporal representations of the driving environment, enabling rollouts of future states~\cite{World_models2018}\add{; under our taxonomy, methods whose principal output is future-scene generation are grouped in Section~\ref{sec5}}. \del{The Dreamer family (DriveDreamer~\cite{DriveDreamer} in Figure~\ref{fig:drive_dreamer}\del{ and ReconDreamer~\cite{ReconDreamer}}) capture compact latent dynamics for long-horizon simulation. Diffusion-based models include DrivingDiffusion~\cite{DrivingDiffusion}, BEVControl~\cite{BEVControl}, and Vista~\cite{Vista}. Transformer-based models such as GAIA-1/2~\cite{GAIA-1} supports multimodal controllable scene generation. Occupancy-based models predict 4D voxelized space, such as OccWorld~\cite{OccWorld}, RenderWorld~\cite{RenderWorld} and DOME~\cite{DOME} combine occupancy forecasting with planning via diffusion. Point-cloud generators like LiDARGen~\cite{LiDARGen} and the unified HERMES~\cite{HERMES} framework produce future LiDAR sweeps or merged 3D representations.}\add{From the embodied-interaction perspective, the more relevant direction is to use such rollouts to improve planning or action generation, as in DrivingGPT~\cite{DrivingGPT}, OccLLaMA~\cite{OccLLaMA}, and Driving in the Occupancy World~\cite{yang2025drivingoccupancyworldvisioncentric}.} \del{Benchmarks in this domain include prediction and occupancy forecasting benchmarks (Occ3D-nuScenes/Waymo~\cite{Occ3D}, OpenScene~\cite{OpenScene}, Precog~\cite{PRECOG}), simulation benchmarks (CarlaSC~\cite{MotionSC}, Bench2Drive~\cite{Bench2Drive}), and end-to-end planning benchmarks (nuPlan~\cite{nuplan}, S2R-Bench~\cite{S2R-Bench}). These collectively test world models across scene generation, motion forecasting, and driving policy integration.}

\del{\noindent\textbf{Hybrid Approaches. }}
Hybrid approaches combine rule-based reliability with learning-based adaptability and cognitive reasoning. Neural waypoint predictors refined by MPC or rule-based controllers augmented with learned forecasting exemplify this paradigm~\cite{Gupta_2023}. Cognitive reasoning extensions include DriveCoT~\cite{DriveCoT}, which integrates Chain-of-Thought supervision into driving tasks, and PRIMEDrive-CoT~\cite{PRIMEDrive-CoT}, which introduces uncertainty-aware reasoning. PlanAgent separates high-level planning from execution, while LeapVAD~\cite{LeapVAD} adopts a dual-process architecture inspired by human cognition. DriveLMM-o1~\cite{DriveLMM-o1} and Reason2Drive~\cite{Reason2Drive} further contribute cognition-augmented datasets for training and evaluating reasoning. 

\noindent The current challenges in autonomous driving are fundamentally tied to the "long tail" problem of rare, high-impact edge cases that are difficult to encounter and address through traditional real-world data collection~\cite{wang2025terasimworldworldwidesafetycriticaldata,ren2025cosmosdrivedreamsscalablesyntheticdriving}. World models solve this by creating a learned, internal simulation of the environment, which functions as a "computational snow globe" where the AI can "mentally rehearse" actions~\cite{Think2Drive,hafner2024masteringdiversedomainsworld}. By generating and training in an unbounded number of synthetic, safety-critical scenarios, these models allow the autonomous system to anticipate and mitigate hazardous situations before they unfold in the physical world, ultimately shifting the paradigm from reactive to predictive autonomy and improving real-world safety~\cite{ding2023surveysafetycriticaldrivingscenario}.

\add{\noindent\textbf{Evaluation Landscape.} }
\add{Evaluation in autonomous driving is necessarily multi-level.}
\add{Rule-based and learning-based planners are commonly tested on CommonRoad, nuPlan, nuScenes, Waymo, CARLA, and AirSim for safety, closed-loop planning, and perception--prediction integration~\cite{CommonRoad,nuplan,nuscenes,Waymo,CARLA,AirSim}.}
\add{Simulation-assisted and hybrid approaches further rely on closed-loop evaluation suites such as Bench2Drive and S2R-Bench to assess planning robustness, controllability, and sim-to-real reliability~\cite{MotionSC,Bench2Drive,S2R-Bench}.}
\add{Together these benchmarks show that progress in driving depends not only on open-loop prediction quality, but also on closed-loop safety under rare and safety-critical scenarios.}

%% file: sec/7_discussion.tex
\section{Discussion}
\label{sec7}

\subsection{\add{Evidence for a Progressive Pathway}}
\label{sec:progression}

\add{Future physical AI will likely integrate four capabilities: perception, reasoning, modeling, and interaction. Rather than a rigid sequence, these capabilities exhibit a structural interdependency where later capabilities rely on structure learned earlier, and existing systems already hint at this progression.}

\add{\noindent\textbf{Positive evidence.} At the \emph{perception $\rightarrow$ reasoning} interface, multimodal systems such as the ICML 2025 physics challenge solution~\cite{Liang2025MultimodalRF} and Physics Supernova~\cite{PhysicsSupernova} extract scene-grounded objects, relations, and diagram structure, then perform multi-step physical inference over them. At the \emph{reasoning $\rightarrow$ modeling} interface, PhysORD~\cite{PhysORD2024}, LagNetViP~\cite{lagnetvip}, and MoSim~\cite{MoSim2025} show that prediction is more stable when rollout dynamics are constrained by explicit physical structure (e.g., Lagrangian priors, conservation relations, symbolic mechanics). At the \emph{modeling $\rightarrow$ interaction} interface, SAIN~\cite{SAIN2019}, X-MOBILITY~\cite{X-MOBILITY}, DrivingGPT~\cite{DrivingGPT}, and OccLLaMA~\cite{OccLLaMA} use predictive latent states or internal rollouts to support control, navigation, and planning. These examples do not prove a universal pipeline, but they suggest that later-stage tasks benefit when earlier-stage capabilities are built in.}

\add{\noindent\textbf{Negative evidence.} Mastering one capability does not generalize easily. On CLEVRER-Humans~\cite{clevrerhumans}, a model strong at \emph{synthetic video perception} and CLEVRER-specific causal QA drops to 54.0\% per-option and 26.9\% per-question accuracy, versus 84.5\% and 71.4\% for humans~\cite{clevrerhumans}.}
Here the missing capability is more human-grounded causal reasoning rather than perception alone. In I-PHYRE~\cite{I-PHYRE}, agents already possess \emph{limited action selection} in simplified physical environments, yet the strongest RL baseline reaches only 57.50\% overall success, compared with 87.55\% for humans~\cite{I-PHYRE}. This suggests that partial interaction competence is still insufficient without stronger predictive modeling for multi-step intervention.
A third failure mode appears in current video generation and world models: they are strong at \emph{perceptual plausibility}, but still underperform on physics benchmarks and real experiments~\cite{Physics-IQ,phygen,worldmodelbench,Morpheus}, indicating that visually convincing modeling does not mean physically faithful thinking, which is needed for robust real-world control in embodied interaction.

\add{Overall, the evidence supports our central thesis: the 4-stage framework is not only a taxonomy, but also a roadmap toward Physical AI in which perception grounds state, reasoning abstracts laws, world modeling projects trajectories, and embodied interaction closes the loop through experience. Although no system yet unifies all four capabilities, existing results increasingly suggest that robust physical intelligence will require their tight integration.}

\del{\noindent\textbf{The Isolation of Perception and Reasoning. }
This survey reveals a fundamental paradox: AI systems achieve superhuman performance on isolated physical tasks, e.g. olympiad-level problem solving, photorealistic video generation, yet lack the flexible, principle-based understanding that allows a child to predict whether stacked blocks will topple. This gap exposes not a quantitative shortfall in data or compute, but a qualitative misalignment between how current systems learn physics and how physical understanding actually works. The four capabilities examined (physical perception, physics reasoning, world modeling and embodied interaction) are not separate research directions but facets of a unified cognitive architecture that current approaches fail to integrate. Physical perception without reasoning remains trapped in correlation; physics reasoning without perceptual grounding produces detached symbolic manipulation; world models without embodied feedback generate plausible but physically inconsistent predictions. Therefore, we advocate that the research community fundamentally rethink how models can mutually inform and constrain these four capabilities through bidirectional coupling, rather than pursuing isolated task improvements.}

\del{\noindent\textbf{Sim-to-real Gap. }}
\add{\subsection{Sim-to-real Gap}}
The transition from simulation to embodied interaction exposes the field's most critical vulnerability. World models trained on internet-scale video generate visually compelling predictions while \del{systematically} violating physical principles, e.g., objects floating without support, collisions without momentum transfer. This fragility stems from optimizing for perceptual plausibility rather than physical consistency. Without embodied consequences to enforce correct physical behavior, systems learn only superficial correlations. We should place greater emphasis on current world models' simulation-to-reality transfer capabilities, for example, by actively refining physical models based on prediction errors encountered during interactions, treating real-world feedback as an essential component of learning rather than merely an end-to-end deployment task. \add{Interaction is not just a deployment target; it can supervise earlier perceptual and predictive stages.}

\del{\noindent\textbf{Internalizing Natural Laws. }}
\add{\subsection{Internalizing Laws of Nature}}
Current approaches reveal a deeper architectural limitation: most systems, including frontier multimodal language models, rely on pattern matching over vast datasets rather than internalizing the compositional, causal structure manifested in the law of nature. This explains why models often perform well on in-distribution benchmarks yet fail catastrophically on counterfactual scenarios or novel configurations. Physics is not a collection of statistical regularities but principles like conservation laws, symmetries, and causal mechanisms that compose systematically. The path forward demands architectures that encode physical law to overcome inductive bias: differentiable physics engines maintaining hard constraints, neuro-symbolic systems integrating learned perception with symbolic reasoning, or embodied learning paradigms acquiring intuition through active intervention rather than passive observation. These architectural innovations provide the possibility for internalizing natural laws. \add{This also helps explain why progress at one stage may not transfer to the next.} We have reason to believe that, beyond the scaling laws of data, parameters, and inference time, a promising path forward may lie in scaling laws grounded in the rules of physical world.

%% file: sec/8_conclusion.tex
\section{Conclusion}
\label{sec8}
\del{This survey presents a task-oriented taxonomy tracing physical AI development through four progressive capabilities}
\add{This survey presents a capability–task hierarchy that traces the development of Physical AI through four progressive stages.}: physical perception, physics reasoning, world modeling, and embodied interaction. Each stage builds upon the previous, advancing from passive observation to active physical comprehension. Despite progress in individual domains, most systems still rely on statistical pattern recognition rather than genuine physical understanding. Moving forward, the most promising path lies in hybrid approaches that integrate physics-grounded architectures, physics-informed training, and symbolic reasoning into unified frameworks. Such frameworks that traverse the entire progression from perception to interaction will enable AI systems to genuinely perceive, reason about, model, and interact with physical reality for safe and reliable deployment.

%% file: sec/x_appendix.tex
\onecolumn
\appendix  

\subsection{\add{PRISMA Process}}
\label{appendix:prisma}

\begin{figure}[t]
\centering
\includegraphics[width=\textwidth]{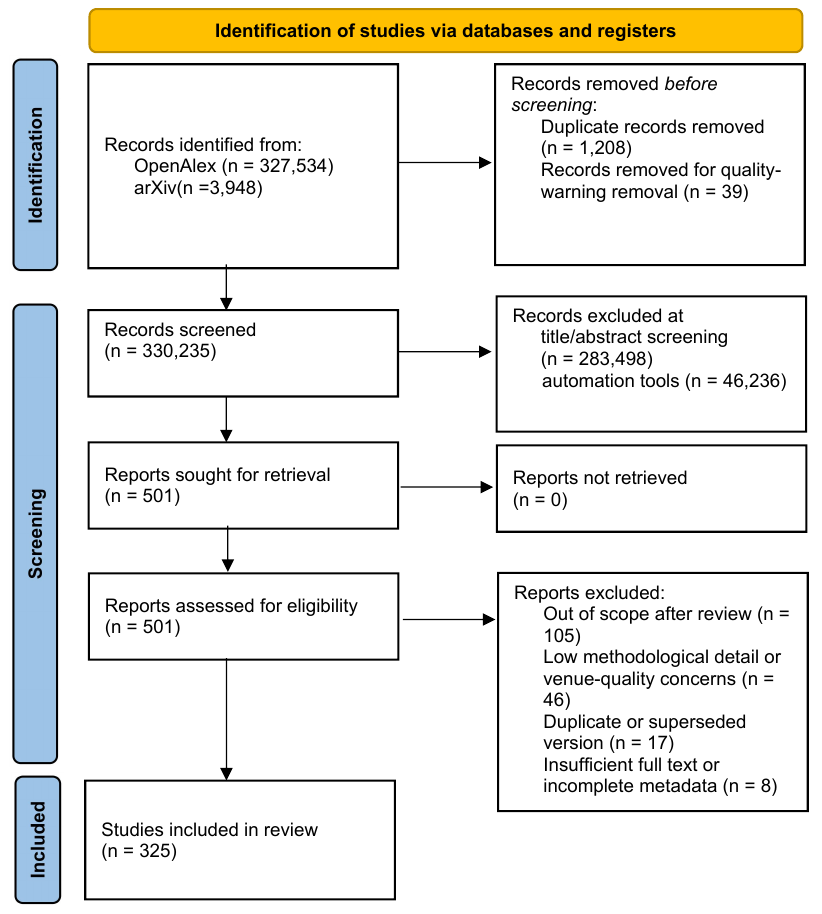}
\caption{\add{PRISMA flow diagram summarizing the literature identification, screening, eligibility assessment, and inclusion process used in this survey.}}
\label{fig:prisma}
\end{figure}

\add{Following the standard PRISMA\footnote{\href{https://www.prisma-statement.org/prisma-2020}{https://www.prisma-statement.org/prisma-2020}} guidelines, Figure~\ref{fig:prisma} explicitly reports the stages of identification, screening, eligibility, and inclusion. We adopt multi-stage pipelines that combine broad database retrieval with title/abstract screening, full-text assessment, and citation-based supplementation to improve coverage. The final set of included studies of 300+ papers was selected from an initial pool of 331,482 records retrieved from OpenAlex\footnote{\href{https://openalex.org/}{https://openalex.org/}} and arXiv, with the intermediate stages involving automated filtering and manual review to ensure relevance, quality, and alignment with our taxonomy. This process is designed to be transparent and reproducible, allowing readers to understand the scope and limitations of our survey while providing a comprehensive overview of the Physical AI landscape.}

\add{\noindent\textbf{Primary Sources and Retrieval Scope.} In our survey, the initial candidate pool was retrieved between January~1,~2012 and March~1,~2026 through venue-based retrieval in OpenAlex together with an explicit arXiv sweep. The OpenAlex query targeted 100+ venues that were identified as relevant to Physical AI based on their historical publication patterns, topical focus, and community recognition. The arXiv sweep covered all papers in the cs.AI, cs.LG, cs.CV, cs.RO, cs.HC, cs.SY, and physics categories to ensure that we captured preprints and emerging work that may not yet be indexed in OpenAlex. This dual-source approach was intended to maximize coverage while allowing for subsequent filtering based on relevance and quality criteria.}

\add{\noindent\textbf{Automated Screening and Preprocessing.} We then applied identifier- and title-level deduplication, with a title-similarity threshold of 0.90, heuristic title/abstract relevance screening, taxonomy-aligned topic matching with a minimum topic score of 0.1, abstract-availability checks except for strong title matches, year-adaptive citation thresholds, and a venue-quality filter that removed 39 blocked-venue records to obtain a focused set of candidate records.  To further refine scope alignment after rule-based preprocessing, we used DeepSeek-V3.2 to screen paper titles, abstracts, and main texts, aiming to retain only papers that matched the capability and task boundaries defined in this survey. To assess the reliability of this LLM-assisted screening stage, we manually audited a sample of 500 records and observed an agreement rate of 88\%, indicating that the model's judgments were sufficiently consistent for large-scale triage. Records that failed these relevance, metadata, or quality gates were excluded before full-text review. For clarity, the ``automation tools'' entry in Figure~\ref{fig:prisma} refers to these deterministic filtering scripts rather than autonomous LLM-only inclusion/exclusion decisions.}

\add{\noindent\textbf{Eligibility Assessment and Manual Verification.} \add{After automated screening, we manually examined the remaining titles, abstracts, and full texts} to determine whether each paper made a substantive methodological, benchmark, or system-level contribution to at least one capability--task node in our taxonomy. Records were excluded when they were out of scope for Physical AI, lacked sufficient technical detail, mainly described products or demos without reproducible methodological content, duplicated a more appropriate archival version, or could not be reliably mapped onto the taxonomy. \add{Borderline cases were resolved by rereading the full text and checking the paper's principal output, evaluation setting, and archival version against the capability--task taxonomy; no paper was included solely on the basis of automated screening. Additional references were incorporated only after manual citation chaining, benchmark-completion passes, or targeted supplementation of official technical reports when they were necessary to document influential systems already discussed in the main text.} Figure~\ref{fig:prisma} therefore makes the intermediate counts and exclusion reasons explicit, rather than reporting only the final set of included studies.}

\subsection{Taxonomy Tables}
\subsubsection{Methodological Papers}
\label{appendix: method}

\input{sec/x_method_tables}

\subsubsection{Benchmark Papers}
\label{appendix: bmk}

\add{The benchmark table below follows the same compact appendix styling as the method tables while keeping the multi-page layout compile-stable.

Website-only products in the Top-1 Model column are cited as footnotes rather than bibliography entries: \webref[Kling official page]{Kling}{https://kling.ai/app/}{Apr. 13, 2026}, \webref[Pika official page]{Pika}{https://pika-art.net/}{Apr. 13, 2026}, and \webref[Pika 2.0 API page]{Pika 2.0}{https://early-access.pika.art/api}{Apr. 13, 2026}. Selected closed-model cards used for Top-1 systems are likewise cited as footnotes: \webref[GPT-4o system card]{GPT-4o}{https://arxiv.org/abs/2410.21276}{Apr. 13, 2026}, \webref[OpenAI o1 system card]{OpenAI o1-mini}{https://arxiv.org/abs/2412.16720}{Apr. 13, 2026}, \webref[Gemini 2.5 Pro model card]{Gemini-2.5-pro}{https://storage.googleapis.com/deepmind-media/Model-Cards/Gemini-2-5-Pro-Model-Card.pdf}{Apr. 13, 2026}, \webref[Gemini 3 Pro model card]{Gemini-3-pro}{https://deepmind.google/models/model-cards/gemini-3-pro/}{Apr. 13, 2026}, \webref[Gemini 3.1 Pro model card]{Gemini-3.1-pro}{https://deepmind.google/models/model-cards/gemini-3-1-pro/}{Apr. 13, 2026}, and \webref[Claude 3.5 Sonnet model card addendum]{Claude-3.5-Sonnet}{https://www.anthropic.com/news/3-5-models-and-computer-use}{Apr. 13, 2026}.}

\input{sec/x_bmk_table}

%% file: sec/x_method_tables.tex

\begingroup
\fontsize{7.0pt}{7.5pt}\selectfont
\setlength{\tabcolsep}{1.2pt}
\renewcommand{\arraystretch}{0.93}
\setlength{\LTleft}{0pt}
\setlength{\LTright}{0pt}
\setlength{\LTpre}{3pt}
\setlength{\LTpost}{3pt}

\begin{longtable}{@{}>{\raggedright\arraybackslash}p{0.23\textwidth}>{\raggedright\arraybackslash}p{0.54\textwidth}>{\centering\arraybackslash}p{0.055\textwidth}>{\raggedright\arraybackslash}p{0.155\textwidth}@{}}
\caption{Methodological papers summarized for physical perception under the current taxonomy.}\\
\toprule
\textbf{Method} & \textbf{Keywords} & \textbf{Year} & \textbf{Venue / DOI} \\
\midrule
\endfirsthead
\caption[]{Methodological papers summarized for physical perception under the current taxonomy. (continued)}\\
\toprule
\textbf{Method} & \textbf{Keywords} & \textbf{Year} & \textbf{Venue / DOI} \\
\midrule
\endhead
\midrule
\multicolumn{4}{r}{\textit{Continued on next page}} \\
\endfoot
\bottomrule
\endlastfoot
\multicolumn{4}{@{}l}{\textbf{Object Recognition}} \\
\midrule
GPT-4V(ision) & MLLM, zero-shot detection, localization, visual grounding & 2023 & GPT-4V system card \\
Mask R-CNN~\cite{he2018maskrcnn} & instance segmentation, detection, region proposals & 2017 & ICCV \\
Grounding DINO~\cite{liu2024groundingdinomarryingdino} & open-set detection, phrase grounding, vision-language pretraining & 2024 & ECCV \\
Qwen3-Omni~\cite{xu2025qwen3omnitechnicalreport} & omni perception, fine-grained recognition, multimodal understanding & 2025 & arXiv:2509.17765 \\
Inter-obj. Graph~\cite{Song_2024} & scene recognition, object relations, graph modeling & 2024 & KBS \\
View-inv. anom. det.~\cite{varghese2025viewinvariantpixelwiseanomalydetection} & pixel anomaly detection, view synthesis, multi-object consistency & 2024 & arXiv:2406.18012 \\
Cog.-guided VAD~\cite{zhang2024cognition} & video anomaly detection, surveillance, cognition priors & 2024 & TSC \\
Expert Ensemble~\cite{Ji_2024} & anomaly ensemble, interaction modeling, driving scenes & 2025 & IJRR \\
\midrule
\multicolumn{4}{@{}l}{\textbf{Spatial Perception}} \\
\midrule
SpIRL~\cite{SpIRL} & spatial relations, contrastive representations, grounding & 2025 & WACV \\
Spatial-MLLM~\cite{wu2025spatialmllmboostingmllmcapabilities} & spatial grounding, multimodal LLMs, spatial IQ & 2025 & arXiv:2505.23747 \\
Multi-SpatialMLLM~\cite{xu2025multi} & multi-frame reasoning, spatial grounding, multimodal LLMs & 2025 & arXiv:2505.17015 \\
ViewSpatial-Bench~\cite{ViewSpatial} & multi-perspective localization, spatial grounding, VLM evaluation & 2025 & arXiv:2505.21500 \\
Multi-view Learning~\cite{zheng2023comprehensive} & multi-view consistency, 3D relations, representation learning & 2023 & Inf. Fusion \\
\midrule
\multicolumn{4}{@{}l}{\textbf{Intrinsic Property Estimation}} \\
\midrule
Intrinsic Img Diff.~\cite{kocsis2024intrinsicimagediffusionindoor} & material estimation, intrinsic images, single-view inference & 2024 & CVPR \\
Fabric Characterization~\cite{hu2020fabricsurfacecharacterizationassessment} & texture representations, fabric analysis, deep features & 2020 & JTI \\
Translucency Prior~\cite{chen2024practical} & translucency estimation, optics, measurement cues & 2024 & CVPR \\
Food Weight Estimation~\cite{Vision-Based} & food weight estimation, monocular inference, visual priors & 2024 & arXiv:2405.16478 \\
Density Recognition~\cite{M_ller_2024} & density estimation, material cues, physical property inference & 2024 & WSCG \\
PhysID~\cite{PhysID} & single-view cues, rigidity inference, physical attributes & 2025 & ICASSP \\
\midrule
\multicolumn{4}{@{}l}{\textbf{Dynamic Estimation}} \\
\midrule
Interaction Network~\cite{interactionnetworks} & object relations, interaction dynamics, graph reasoning & 2016 & NIPS \\
Visual Interaction Network~\cite{VIN} & video dynamics, object-centric simulation, future-state prediction & 2017 & NIPS \\
Neural Physics Engine~\cite{NPE} & compositional dynamics, object-based physics, simulation & 2017 & ICLR \\
Newtonian~\cite{NewtonianIU} & static-image dynamics, motion inference, Newtonian scenarios & 2016 & CVPR \\
PhysID~\cite{PhysID} & single-view dynamics, interaction cues, rigid-body inference & 2025 & ICASSP \\
SlotFormer~\cite{SlotFormer} & object-centric dynamics, unsupervised simulation, video modeling & 2023 & ICLR \\
Reason.-enh. OCL~\cite{li2025reasoningenhancedobjectcentriclearningvideos} & object-centric learning, video reasoning, interaction structure & 2025 & KDD \\
\end{longtable}
\vspace{0.5em}

\begin{longtable}{@{}>{\raggedright\arraybackslash}p{0.23\textwidth}>{\raggedright\arraybackslash}p{0.54\textwidth}>{\centering\arraybackslash}p{0.055\textwidth}>{\raggedright\arraybackslash}p{0.155\textwidth}@{}}
\caption{Methodological papers summarized for physics reasoning under the current taxonomy.}\\
\toprule
\textbf{Method} & \textbf{Keywords} & \textbf{Year} & \textbf{Venue / DOI} \\
\midrule
\endfirsthead
\caption[]{Methodological papers summarized for physics reasoning under the current taxonomy. (continued)}\\
\toprule
\textbf{Method} & \textbf{Keywords} & \textbf{Year} & \textbf{Venue / DOI} \\
\midrule
\endhead
\midrule
\multicolumn{4}{r}{\textit{Continued on next page}} \\
\endfoot
\bottomrule
\endlastfoot
\multicolumn{4}{@{}l}{\textbf{Symbolic Reasoning}} \\
\midrule  
CoT~\cite{CoT} & multi-step prompting, derivation scaffolds, symbolic reasoning & 2022 & NeurIPS \\
Phys. Supernova~\cite{PhysicsSupernova} & tool-augmented agents, olympiad problems, long-horizon reasoning & 2025 & arXiv:2509.01659 \\
LOCA-R~\cite{Jian2025LOCA} & local reasoning, verifiable steps, olympiad physics & 2025 & arXiv:2511.10515 \\
Intro-phys. Agent~\cite{Kortemeyer2023CouldAAF} & textbook QA, physics QA, course-level problem solving & 2023 & PRPER \\
Steps~\cite{Addala2024StepsAAA} & prompting, STEM reasoning, problem decomposition & 2024 & arXiv:2412.05023 \\
Symbolic or Numerical?~\cite{Dan2025SymbolicOND} & symbolic versus numerical reasoning, physics QA, error analysis & 2025 & arXiv:2507.01334 \\
KG for Physics QA~\cite{Addala2024KnowledgeGAE} & knowledge graphs, QA decomposition, retrieval & 2024 & arXiv:2412.05453 \\
LLMPhy~\cite{LLMPhy_2025} & LLM plus world models, latent simulation, complex physics & 2024 & arXiv:2411.08027 \\
\midrule
\multicolumn{4}{@{}l}{\textbf{Multimodal-grounded Reasoning}} \\
\midrule  
DCL~\cite{chen2021grounding} & video grounding, object-event relations, dynamic reasoning & 2021 & ICLR \\
TRACE~\cite{Imani2025TRACE} & stepwise reasoning, grounding errors, VLM analysis & 2025 & EACL \\
MR-Science~\cite{Liang2025MultimodalRF} & diagram understanding, vision-text integration, science reasoning & 2025 & arXiv:2509.06079 \\
P1-VL~\cite{luo2026p1} & perception-reasoning coupling, olympiad physics, multimodal VLMs & 2026 & arXiv:2602.09443 \\
\midrule
\multicolumn{4}{@{}l}{\textbf{Causal and Counterfactual Reasoning}} \\
\midrule  
Causal Dynamical System~\cite{causalmodeling} & dynamical systems, causal structure, latent dependencies & 2018 & arXiv:1803.08784 \\
Neural Causal Model~\cite{ke2020causalmodels} & unknown interventions, causal discovery, counterfactual inference & 2019 & arXiv:1910.01075 \\
Causal Threads~\cite{causalthreads} & causal traces, interpretable changes, dynamic systems & 2023 & ICADL \\
Diff-physics reasoning~\cite{ding2021dynamic} & differentiable physics, vision-language reasoning, counterfactuals & 2021 & NeurIPS \\
\midrule
\multicolumn{4}{@{}l}{\textbf{Accelerate Physics Research}} \\
\midrule  
SRBench++~\cite{deFranca2025SRBenchPP} & symbolic regression, domain-expert interpretation, physical laws & 2025 & TEVC \\
GA-SISSO~\cite{mazheika2024_ga_sisso} & feature selection, symbolic regression, compressed sensing & 2024 & arXiv:2403.15816 \\
Domain-aware SR~\cite{huang2025_domain_aware_sr_priors} & symbolic priors, domain constraints, expression parsing & 2025 & arXiv:2503.09592 \\
ASP-assisted SR~\cite{aravanis2025_asp_sr_fluid} & symbolic regression, fluid mechanics, hidden physics & 2025 & arXiv:2507.17777 \\
Unit-constrained SR~\cite{zhang2024_interpretable_turbulence_sr} & symbolic regression, unit consistency, turbulence modeling & 2025 & Acta Mech. \\
The AI Scientist~\cite{lu2024aiscientist} & autonomous discovery, research agents, open-ended workflows & 2024 & arXiv:2408.06292 \\
The AI Scientist-v2~\cite{yamada2024aiscientistv2} & tree search, autonomous discovery, workshop-level research & 2025 & arXiv:2504.08066 \\
CMBAgent~\cite{Xu2025cmbagent} & planning and control, language agents, autonomous discovery & 2025 & ICML Workshop \\
AtomAgents~\cite{ghafarollahi2024atomagents} & materials discovery, multi-agent collaboration, physics-aware search & 2024 & arXiv:2407.10022 \\
Materials-law Disc.~\cite{hu2024_multiagent_materials_discovery} & materials laws, multi-agent discovery, scientific law induction & 2024 & arXiv:2411.16416 \\
Turbulence Modeling~\cite{yang2025_llm_turbulence_modeling} & turbulence modeling, scientific modeling, LLM assistance & 2025 & Flow \\
ScienceClaw~\cite{Wang2026ScienceClaw} & distributed discovery, artifact exchange, multi-agent systems & 2026 & arXiv:2603.14312 \\
ClawdLab / BeachSci~\cite{Weidener2026ClawdLab} & OpenClaw ecosystem, autonomous research, agent coordination & 2026 & arXiv:2602.19810 \\
\end{longtable}
\vspace{0.5em}

\begin{longtable}{@{}>{\raggedright\arraybackslash}p{0.23\textwidth}>{\raggedright\arraybackslash}p{0.54\textwidth}>{\centering\arraybackslash}p{0.055\textwidth}>{\raggedright\arraybackslash}p{0.155\textwidth}@{}}
\caption{Methodological papers summarized for world modeling under the current taxonomy.}\\
\toprule
\textbf{Method} & \textbf{Keywords} & \textbf{Year} & \textbf{Venue / DOI} \\
\midrule
\endfirsthead
\caption[]{Methodological papers summarized for world modeling under the current taxonomy. (continued)}\\
\toprule
\textbf{Method} & \textbf{Keywords} & \textbf{Year} & \textbf{Venue / DOI} \\
\midrule
\endhead
\midrule
\multicolumn{4}{r}{\textit{Continued on next page}} \\
\endfoot
\bottomrule
\endlastfoot
\multicolumn{4}{@{}l}{\textbf{Image Generation}} \\
\cmidrule(lr){1-4}
Phong shading~\cite{Phong} & illumination models, shading, computer graphics & 1975 & CACM \\
Cook-Torrance~\cite{Cook-Torrance} & reflectance modeling, physically based rendering, materials & 1982 & TOG \\
Orbit / Isaac Sim~\cite{isaacsim} & simulation framework, photorealism, robot scenes & 2023 & RA-L \\
Unity ML-Agents~\cite{Unity-ML-Agents} & simulation platform, environment generation, embodied learning & 2018 & arXiv:1809.02627 \\
Stereo Sensor Simulation~\cite{stereovision-depth-sensors} & sensor simulation, active stereo, sim-to-real transfer & 2023 & T-RO \\
Ref-NeRF~\cite{Ref-NeRF} & neural rendering, view-dependent appearance, NeRFs & 2022 & CVPR \\
ENVIDR~\cite{ENVIDR} & differentiable rendering, environment lighting, neural rendering & 2023 & ICCV \\
\midrule
\multicolumn{4}{@{}l}{\textbf{Video Generation}} \\
\cmidrule(lr){1-4}
GPT4Motion~\cite{GPT4Motion} & text-to-video generation, physics planning, Blender simulation & 2024 & CVPRW \\
MotionCraft~\cite{MotionCraft} & physics-based generation, zero-shot video synthesis, motion priors & 2024 & NeurIPS \\
VideoREPA~\cite{VideoREPA} & relational alignment, physics learning, video generation & 2025 & NeurIPS \\
Open-Sora~\cite{opensora} & open-source video generation, diffusion, world models & 2024 & arXiv:2412.20404 \\
CogVideoX~\cite{CogVideoX} & text-to-video diffusion, expert transfer, long videos & 2025 & ICLR \\
LaVie~\cite{wang2023lavie} & latent diffusion, cascaded generation, video synthesis & 2024 & IJCV \\
VLIPP~\cite{VLIPP} & physics priors, plausible video generation, VLM guidance & 2025 & ICCV \\
PhysGen~\cite{PhysGen} & rigid-body grounding, image-to-video, physics-aware generation & 2024 & ECCV \\
ProPhy~\cite{Wang2025ProPhyPP} & progressive physical alignment, dynamic simulation, video generation & 2025 & arXiv:2512.05564 \\
DiffPhy~\cite{DiffPhy} & LLM-guided physics, video generation, reasoning priors & 2025 & arXiv:2505.21653 \\
DINO-Foresight~\cite{Karypidis2024DINOForesightLI} & future prediction, representation forecasting, self-supervision & 2025 & NeurIPS \\
GAIA-1~\cite{GAIA-1} & generative world model, autonomous driving, controllable synthesis & 2023 & arXiv:2309.17080 \\
\midrule
\multicolumn{4}{@{}l}{\textbf{Scene Reconstruction}} \\
\cmidrule(lr){1-4}
SfM~\cite{SFM} & geometry recovery, motion cues, 3D reconstruction & 1979 & Proc. R. Soc. B \\
Multi-view Stereo~\cite{MVS} & multi-view depth, 3D geometry, scene reconstruction & 2006 & CVPR \\
PhyRecon~\cite{PhyRecon} & neural scene reconstruction, physical plausibility, scene priors & 2024 & NeurIPS \\
IDR~\cite{IDR} & neural surface reconstruction, geometry-appearance disentanglement & 2020 & NeurIPS \\
ReconDreamer~\cite{ReconDreamer} & driving reconstruction, restoration, world models & 2025 & CVPR \\
DriveDreamer4D~\cite{DriveDreamer4D} & 4D scene representations, driving world models, reconstruction & 2025 & CVPR \\
\midrule
\multicolumn{4}{@{}l}{\textbf{Physics-constrained Simulation}} \\
\cmidrule(lr){1-4}
DINO-world~\cite{DINO-world2025} & video world models, foundation features, latent dynamics & 2025 & arXiv:2507.19468 \\
MuJoCo~\cite{MuJoCo} & physics engine, model-based control, differentiable simulation support & 2012 & IROS \\
Brax~\cite{Brax} & differentiable physics, rigid-body simulation, control & 2021 & NeurIPS \\
NeoPhysIx~\cite{NeoPhysIx} & fast 3D simulation, AI development, physics simulation & 2024 & arXiv:2411.05799 \\
PhysORD~\cite{PhysORD2024} & Lagrangian constraints, vehicle dynamics, physics-embedded prediction & 2025 & arXiv:2503.06748 \\
Object Nets~\cite{SAIN2019} & object-based simulation, control, physics priors & 2019 & ICRA \\
DEM-NeRF~\cite{tan2025_dem_nerf_neurosymbolic} & neuro-symbolic simulation, physics-informed discovery, scientific modeling & 2025 & arXiv:2507.21350 \\
MoSim~\cite{MoSim2025} & motion simulation, RL world models, neural simulation & 2025 & CVPR \\
LagNetViP~\cite{lagnetvip} & Lagrangian dynamics, video prediction, inductive biases & 2020 & AAAI \\
FusionForce~\cite{fusionforce2025} & trajectory prediction, neuro-symbolic fusion, differentiable layers & 2025 & arXiv:2502.10156 \\
\end{longtable}
\vspace{0.5em}

\begin{longtable}{@{}>{\raggedright\arraybackslash}p{0.23\textwidth}>{\raggedright\arraybackslash}p{0.54\textwidth}>{\centering\arraybackslash}p{0.055\textwidth}>{\raggedright\arraybackslash}p{0.155\textwidth}@{}}
\caption{Methodological papers summarized for embodied interaction under the current taxonomy.}\\
\toprule
\textbf{Method} & \textbf{Keywords} & \textbf{Year} & \textbf{Venue / DOI} \\
\midrule
\endfirsthead
\caption[]{Methodological papers summarized for embodied interaction under the current taxonomy. (continued)}\\
\toprule
\textbf{Method} & \textbf{Keywords} & \textbf{Year} & \textbf{Venue / DOI} \\
\midrule
\endhead
\midrule
\multicolumn{4}{r}{\textit{Continued on next page}} \\
\endfoot
\bottomrule
\endlastfoot
\multicolumn{4}{@{}l}{\textbf{Robotics}} \\
\cmidrule(lr){1-4}
Gato~\cite{ReedEtAl2022_Gato} & generalist agents, sequence modeling, multi-domain control & 2022 & TMLR \\
RT-1~\cite{BrohanEtAl2022_RT1} & robot transfer, real-world control, large-scale data & 2023 & RSS \\
RT-2~\cite{Brohan2023RT2VM} & vision-language-action models, web transfer, robotic control & 2023 & CoRL \\
$\pi_0$~\cite{BlackEtAl2024_pi0} & flow matching, robot control, vision-language-action & 2025 & arXiv:2410.24164 \\
OpenVLA~\cite{KimEtAl2024_OpenVLA} & open-source vision-language-action, control, action tokens & 2025 & CoRL \\
Gemini Robotics & real-world grounding, dexterous control, multi-embodiment & 2025 & Gemini Robotics Blog \\
Safe Control~\cite{Safety-Critical} & continuous control, safety constraints, robotics & 2021 & IEEE Ctrl. Syst. Lett. \\
Flow Matching~\cite{lipman2023_flow_matching} & continuous generative control, trajectory matching, smooth actions & 2023 & ICLR \\
Diffusion Policy~\cite{ChiEtAl2023_DiffusionPolicy} & action diffusion, visuomotor policies, continuous control & 2024 & IJRR \\
Open X / RT-X~\cite{BrohanEtAl2022_OpenX} & cross-platform generalization, heterogeneous data, robot transfer & 2024 & ICRA \\
World Action Models~\cite{Ye2026WorldAM} & future-state prediction, zero-shot policies, embodied control & 2026 & arXiv:2602.15922 \\
World Action Verifier~\cite{Liu2026WorldAV} & forward-inverse asymmetry, self-improving world models, verification & 2026 & arXiv:2604.01985 \\
DriveVA~\cite{Liu2026DriveVAVA} & video action models, zero-shot driving, action generation & 2026 & arXiv:2604.04198 \\
Active Exploration~\cite{LingEtAl2023_ActiveLearning} & online adaptation, exploration, contact-rich manipulation & 2023 & ICRA \\
Affordance Control~\cite{SeoEtAl2023_HybridControl} & language grounding, affordances, hybrid control & 2022 & CoRL \\
Active Visuo-tactile~\cite{WuEtAl2023_ObjectProperty} & visuo-tactile sensing, property-aware manipulation, tactile grounding & 2023 & arXiv:2310.15551 \\
\midrule
\multicolumn{4}{@{}l}{\textbf{Navigation}} \\
\cmidrule(lr){1-4}
GOSE~\cite{chaplot2020objectgoalnavigationusing} & semantic exploration, object navigation, mapping & 2020 & NeurIPS \\
RIM~\cite{RIM} & implicit maps, object navigation, recursive mapping & 2023 & IROS \\
LOAT~\cite{LOTA} & LLM object affinities, zero-shot object navigation, transfer & 2024 & arXiv:2403.09971 \\
LGR~\cite{LGR} & frontier ranking, LLM guidance, object navigation & 2025 & arXiv:2503.20241 \\
CL-CoTNav~\cite{CL-CoTNav} & closed-loop chain-of-thought, zero-shot object navigation, VLMs & 2025 & arXiv:2504.09000 \\
ASCENT~\cite{ASCENT} & floor-aware exploration, coarse-to-fine search, object navigation & 2025 & arXiv:2505.23019 \\
BabyWalk~\cite{BabyWalk} & step decomposition, long-horizon vision-language navigation, curriculum & 2020 & ACL \\
ADAPT~\cite{ADAPT} & modality-aligned prompts, vision-language navigation, action prompting & 2022 & CVPR \\
HAMT~\cite{HAMT} & history-aware transformers, vision-language navigation, multimodal memory & 2021 & NeurIPS \\
EnvDrop~\cite{EnvDrop} & back-translation, unseen environments, VLN generalization & 2019 & NAACL \\
DUET~\cite{DUET} & dual-scale graph transformers, global-local modeling, VLN & 2022 & CVPR \\
NavGPT~\cite{NavGPT} & explicit reasoning, vision-language navigation, LLM planning & 2024 & AAAI \\
NavCoT~\cite{NavCoT} & disentangled reasoning, vision-language navigation, chain-of-thought & 2025 & TPAMI \\
VELMA~\cite{VELMA} & street-view navigation, verbalized reasoning, VLMs & 2024 & AAAI \\
NaVILA~\cite{NaVILA} & legged navigation, vision-language-action, locomotion & 2025 & RSS \\
CMN~\cite{CMN} & cross-modal memory, dialog navigation, ambiguity resolution & 2020 & CVPR \\
GVDN~\cite{GVDN} & goal-oriented dialog navigation, reinforcement learning, dialog grounding & 2022 & EMNLP Find. \\
FLAME~\cite{FLAME} & urban navigation, multimodal LLMs, outdoor reasoning & 2025 & AAAI \\
X-MOBILITY~\cite{X-MOBILITY} & world models, end-to-end navigation, sim-to-real transfer & 2025 & ICRA \\
TWIST~\cite{TWIST} & world-model distillation, teacher-student transfer, sim-to-real & 2024 & ICRA \\
\midrule
\multicolumn{4}{@{}l}{\textbf{Autonomous Driving}} \\
\cmidrule(lr){1-4}
E2E Self-driving~\cite{bojarski2016endendlearningselfdriving} & imitation learning, steering prediction, perception-to-control & 2016 & arXiv:1604.07316 \\
Intelligent Driver Model~\cite{IDM} & car-following, rule-based control, traffic flow & 2000 & Phys. Rev. E \\
RRT~\cite{lavalle1998rapidly} & sampling-based planning, path planning, motion trees & 1998 & INRIA RR 9811 \\
MPC~\cite{MPC} & trajectory optimization, planning, receding-horizon control & 2003 & Control Eng. Pract. \\
Potential Fields~\cite{khatib1986real} & obstacle avoidance, reactive planning, mobile robots & 1986 & IJRR \\
ChauffeurNet~\cite{ChauffeurNet} & imitation learning, robust driving, data synthesis & 2019 & RSS \\
MaxEnt IRL~\cite{Maximum} & inverse reinforcement learning, reward inference, driving policy & 2008 & AAAI \\
LaneGCN~\cite{LaneGCN} & lane-graph forecasting, trajectory prediction, motion reasoning & 2020 & ECCV \\
Wayformer~\cite{Wayformer} & attention forecasting, multimodal prediction, efficiency & 2022 & arXiv:2207.05844 \\
AlphaDrive~\cite{AlphaDrive} & vision-language models plus RL, reasoning, autonomous driving & 2025 & arXiv:2503.07608 \\
AutoDrive-R$^2$~\cite{AutoDrive} & self-reflection, VLA driving, reasoning with RL & 2025 & arXiv:2509.01944 \\
World Models~\cite{World_models2018} & latent world models, imagination, planning & 2018 & arXiv:1803.10122 \\
DrivingGPT~\cite{DrivingGPT} & world models plus planning, multimodal autoregression, driving & 2025 & ICCV \\
OccLLaMA~\cite{OccLLaMA} & occupancy-language-action, world models, planning & 2024 & arXiv:2409.03272 \\
Occ. World Driving~\cite{yang2025drivingoccupancyworldvisioncentric} & 4D occupancy, planning, world models & 2025 & AAAI \\
DriveDreamer~\cite{DriveDreamer} & real-world world models, scene generation, driving simulation & 2024 & ECCV \\
IA-MPC~\cite{Gupta_2023} & hybrid control, interaction-aware planning, MPC & 2023 & ICRA \\
DriveCoT~\cite{DriveCoT} & chain-of-thought, end-to-end driving, reasoning supervision & 2024 & arXiv:2403.16996 \\
PRIMEDrive-CoT~\cite{PRIMEDrive-CoT} & uncertainty-aware chain-of-thought, object interactions, driving & 2025 & CVPRW \\
LeapVAD~\cite{LeapVAD} & dual-process cognition, perception, driving & 2025 & TNNLS \\
DriveLMM-o1~\cite{DriveLMM-o1} & multimodal driving reasoning, stepwise supervision, scenarios & 2025 & arXiv:2503.10621 \\
Reason2Drive~\cite{Reason2Drive} & interpretable reasoning, driving, multimodal understanding & 2024 & ECCV \\
TeraSim-World~\cite{wang2025terasimworldworldwidesafetycriticaldata} & safety-critical synthesis, synthetic driving data, long-tail scenarios & 2025 & arXiv:2509.13164 \\
Cosmos-Drive-Dreams~\cite{ren2025cosmosdrivedreamsscalablesyntheticdriving} & world foundation models, synthetic driving data, scalable generation & 2025 & arXiv:2506.09042 \\
Think2Drive~\cite{Think2Drive} & latent world-model reinforcement learning, efficient driving, CARLA & 2024 & ECCV \\
DreamerV3~\cite{hafner2024masteringdiversedomainsworld} & world models, control generation, latent imagination & 2025 & Nature \\
\end{longtable}
\endgroup

%% file: sec/x_bmk_table.tex
\begingroup
\fontsize{7.0pt}{7.5pt}\selectfont
\setlength{\tabcolsep}{1.2pt}
\renewcommand{\arraystretch}{0.93}
\setlength{\LTleft}{0pt}
\setlength{\LTright}{0pt}
\setlength{\LTpre}{3pt}
\setlength{\LTpost}{3pt}
\begin{xltabular}{\textwidth}{@{}
>{\raggedright\arraybackslash}p{0.17\textwidth}
>{\raggedright\arraybackslash}p{0.15\textwidth}
>{\raggedright\arraybackslash}p{0.14\textwidth}
>{\raggedright\arraybackslash}p{0.22\textwidth}
>{\raggedright\arraybackslash}p{0.24\textwidth}
>{\centering\arraybackslash}p{0.06\textwidth}
@{}}
\caption{Examples of benchmark summary entries used throughout the survey.}
\label{tab:benchmark_summary_template} \\
\toprule
\textbf{Benchmark} & \textbf{Metrics} & \textbf{Modality} & \textbf{Task} & \textbf{Top-1 Model} & \textbf{Year} \\
\midrule
\endfirsthead
\caption[]{Examples of benchmark summary entries used throughout the survey. (continued)} \\
\toprule
\textbf{Benchmark} & \textbf{Metrics} & \textbf{Modality} & \textbf{Task} & \textbf{Top-1 Model} & \textbf{Year} \\
\midrule
\endhead
\midrule
\multicolumn{6}{r}{\textit{Continued on next page}} \\
\endfoot
\bottomrule
\endlastfoot

\multicolumn{6}{@{}l}{\textbf{Physical Perception}} \\
\midrule
ImageNet~\cite{Imagenet} 
& Acc. 
& Image 
& \makecell[l]{Object Recognition,\\hierarchy tagging} 
& \makecell[l]{/} 
& 2009 \\

COCO~\cite{coco} 
& mAP
& Text + Image
& \makecell[l]{Object Recognition,\\context segmentation} 
& \makecell[l]{/ } 
& 2014 \\

BOP~\cite{BOP} 
& AR \& AP
& Image 
& \makecell[l]{Object Recognition,\\geometry-based 6D}
& \makecell[l]{Vidal-18~\cite{vidal-18}} 
& 2018 \\

Phys-AD~\cite{Phys-AD} 
& AUROC \& PAEval 
& Video
& \makecell[l]{Object Recognition,\\interactive abnormality} 
& \makecell[l]{MNAD~\cite{MNAD}} 
& 2025 \\

SpatialScore~\cite{SpatialScore} 
& Acc.
& Text + Image 
& \makecell[l]{Spatial Perception,\\multimodal QA} 
& \makecell[l]{Gemini-3-pro} 
& 2025 \\

ViewSpatial-Bench~\cite{ViewSpatial}
& Acc.
& Text + Image
& \makecell[l]{Spatial Perception,\\multi-perspective localization}
& \makecell[l]{/}
& 2025 \\

Open3DVQA~\cite{Open3DVQA} 
& Acc.
& Text + Image 
& \makecell[l]{Spatial Perception,\\spatial VQA} 
& \makecell[l]{Qwen2-VL-7B(FT)~\cite{qwen2vl}} 
& 2025 \\

MMSI-Bench~\cite{yang2025mmsi} 
& Acc.
& Text + Image 
& \makecell[l]{Spatial Perception,\\multi-image VQA} 
& \makecell[l]{Gemini-3-pro} 
& 2025 \\

Physion~\cite{Physion}
& Acc. 
& Video 
& \makecell[l]{Dynamic Estimation,\\outcome prediction} 
& \makecell[l]{DPI-Net~\cite{dpi-net}} 
& 2021 \\

Physion++~\cite{PhysionPlus}
& Acc.
& Video
& \makecell[l]{Dynamic Estimation,\\property inference} 
& \makecell[l]{DPI-Net~\cite{dpi-net}} 
& 2023 \\

I-PHYRE~\cite{I-PHYRE}
& Reward SR
& Interactive Environment
& \makecell[l]{Dynamic Estimation,\\softbody reasoning} 
& \makecell[l]{/} 
& 2023 \\

ContPhy~\cite{ContPhy}
& Acc.
& Video
& \makecell[l]{Dynamic Estimation,\\interactive reasoning} 
& \makecell[l]{ContPRO~\cite{ContPhy}} 
& 2024 \\

\midrule
\multicolumn{6}{@{}l}{\textbf{Physics Reasoning}} \\
\midrule

PhysicsEval~\cite{physicseval}
& PPS Score
& Text
& \makecell[l]{Symbolic Reasoning,\\technique evaluation} 
& \makecell[l]{Phi-4-reasoning-plus\\ (Multi-Agent)~\cite{phi4reasoning}} 
& 2025 \\

UGPhysics~\cite{ugphysics}
& Acc.
& Text
& \makecell[l]{Symbolic Reasoning,\\ textbook-style QA} 
& \makecell[l]{OpenAI-o1-mini} 
& 2025 \\

PHYBench~\cite{qiu2025phybench}
& Acc. \& EED Score 
& Text
& \makecell[l]{Symbolic Reasoning,\\physics expression} 
& \makecell[l]{Gemini-2.5-pro} 
& 2025 \\

ABench-Physics~\cite{abenchphysics}
& Acc.
& Text
& \makecell[l]{Symbolic Reasoning,\\numerical QA} 
& \makecell[l]{Gemini-2.5-pro} 
& 2025 \\

GPQA~\cite{GPQA}
& Acc.
& Text
& \makecell[l]{Symbolic Reasoning,\\expert QA} 
& \makecell[l]{Gemini-3.1-pro} 
& 2023 \\

OlympiadBench~\cite{olympiadbench}
& Acc.
& Text + Image
& \makecell[l]{Symbolic Reasoning,\\olympiad-level QA}
& \makecell[l]{/}
& 2024 \\

PhysReason~\cite{physreason}
& Acc.
& Text + Image
& \makecell[l]{Multimodal Reasoning,\\process evaluation} 
& \makecell[l]{Deepseek-R1~\cite{guo2025deepseek}} 
& 2025 \\

CLEVRER~\cite{clevrer}
& Acc.
& Text + Video
& \makecell[l]{Multimodal Reasoning,\\video causality} 
& \makecell[l]{NS-DR~\cite{clevrer}} 
& 2020 \\

ComPhy~\cite{ComPhy}
& Acc.
& Text + Video
& \makecell[l]{Multimodal Reasoning,\\counterfactual prediction} 
& \makecell[l]{CPL~\cite{ComPhy}} 
& 2022 \\

MMMU~\cite{mmmu}
& Acc.
& Text + Image
& \makecell[l]{Multimodal Reasoning,\\discipline-diverse QA} 
& \makecell[l]{GPT-4o} 
& 2024 \\

MMMU-Pro~\cite{mmmupro}
& Acc.
& Text + Image
& \makecell[l]{Multimodal Reasoning,\\vision-only QA} 
& \makecell[l]{GPT-4o} 
& 2024 \\

SeePhys~\cite{seephys}
& Acc.
& Text + Image
& \makecell[l]{Multimodal Reasoning,\\diagram-based QA} 
& \makecell[l]{Gemini-2.5-pro} 
& 2025 \\

CAUSAL3D~\cite{causal3D}
& Acc. \& F1 Score
& Text + Image
& \makecell[l]{Causal Reasoning,\\ structure learning} 
& \makecell[l]{/} 
& 2025 \\

CLEVRER-Humans~\cite{clevrerhumans}
& Acc.
& Text + Video
& \makecell[l]{Counterfactual Reasoning,\\ human-centric QA} 
& \makecell[l]{ALOE~\cite{ALOE}} 
& 2022 \\

PhySense~\cite{physense}
& Acc.
& Text + Image
& \makecell[l]{Causal and Counterfactual\\ Reasoning, principle-first QA} 
& \makecell[l]{Gemini-2.5-pro} 
& 2025 \\

SRBench~\cite{LaCava2021SRBench}
& R$^2$ \& recovery metrics
& Tabular + Equations
& \makecell[l]{Accelerate Physics Research,\\symbolic regression}
& \makecell[l]{/}
& 2021 \\

SRBench++~\cite{deFranca2025SRBenchPP}
& Interpretability \& recovery metrics
& Tabular + Equations
& \makecell[l]{Accelerate Physics Research,\\physics-aware symbolic regression}
& \makecell[l]{/}
& 2025 \\

ScienceAgent Bench~\cite{scienceagentbench}
& VER \& SR \& CBS \& Cost
& Text
& \makecell[l]{Accelerate Physics Research,\\ data-driven discovery} 
& \makecell[l]{Claude-3.5-Sonnet} 
& 2024 \\

SciCode~\cite{scicode}
& Pass@1
& Text
& \makecell[l]{Accelerate Physics Research,\\scientific programming} 
& \makecell[l]{Claude-3.5-Sonnet} 
& 2024 \\

\midrule
\multicolumn{6}{@{}l}{\textbf{World Modeling}} \\
\midrule

NYU Depth V2~\cite{NYU-Depth-V2}
& Acc.
& Text + Image
& \makecell[l]{Image Generation,\\scene parsing} 
& \makecell[l]{/} 
& 2012 \\

PhyGen~\cite{phygen}
& PhyGenEval
& T2V
& \makecell[l]{Video Generation,\\intuitive physics} 
& \makecell[l]{Kling} 
& 2024 \\

WorldModel Bench~\cite{worldmodelbench}
& Instruction Following \& Physics Adherence \& Commonsense
& T2V \& I2V
& \makecell[l]{Video Generation,\\dynamic modeling} 
& \makecell[l]{Kling} 
& 2025 \\

VideoPhy~\cite{2024videophy}
& SA \& PC
& T2V
& \makecell[l]{Video Generation,\\ physical plausibility} 
& \makecell[l]{Pika} 
& 2024 \\

PhysicsIQ~\cite{Physics-IQ}
& Physics-IQ Score 
& Video
& \makecell[l]{Video Generation,\\world simulation} 
& \makecell[l]{VideoPoet~\cite{VideoPoet}} 
& 2025 \\

Morpheus~\cite{Morpheus}
& Physical-reasoning score
& Video
& \makecell[l]{Video Generation,\\real physical experiments}
& \makecell[l]{/}
& 2025 \\

DTU multi-view stereo~\cite{DTU2014}
& Acc. \& Completeness
& Image
& \makecell[l]{Scene Reconstruction,\\stereo evaluation} 
& \makecell[l]{/} 
& 2014 \\

Tanks and Temples~\cite{TanksAndTemples2017}
& Precision \& Recall
& Image
& \makecell[l]{Scene Reconstruction,\\large-scale evaluation} 
& \makecell[l]{/} 
& 2017 \\

ETH3D~\cite{ETH3D2017}
& Accuracy \& Completeness \& F1 score
& Image + Video
& \makecell[l]{Scene Reconstruction,\\high-resolution stereo} 
& \makecell[l]{/} 
& 2017 \\

DriverDreamer 4D~\cite{DriveDreamer4D}
& NTA-IoU \& NTL-IoU \& FID
& Image + LiDAR
& \makecell[l]{Scene Reconstruction,\\} 
& \makecell[l]{DriveDreamer4D~\cite{DriveDreamer4D}} 
& 2024 \\

PHYRE~\cite{phyre}
& AUCCESS
& Image
& \makecell[l]{Physics Simulation,\\goal-driven puzzle} 
& \makecell[l]{DQN-O~\cite{phyre}} 
& 2019 \\

PhyWorld-Bench~\cite{PhyWorldBench}
& SA \& PC
& T2V
& \makecell[l]{Physics Simulation,\\physical realism} 
& \makecell[l]{Pika 2.0} 
& 2025 \\

PhyBlock~\cite{Ma2025PhyBlockAP}
& SR
& 3D Blocks + Actions
& \makecell[l]{Physics Simulation,\\block assembly planning}
& \makecell[l]{/}
& 2025 \\

\midrule
\multicolumn{6}{@{}l}{\textbf{Embodied Interaction}} \\
\midrule

EmbodiedBench~\cite{EmbodiedBench}
& SR
& Text + Image
& \makecell[l]{Robotics,\\vision-driven agents} 
& \makecell[l]{GPT-4o} 
& 2025 \\

EMMOE~\cite{EMMOE}
& SR
& Text + Image
& \makecell[l]{Robotics,\\ mobile manipulation} 
& \makecell[l]{HOMIEBOT~\cite{EMMOE}} 
& 2025 \\

Open X-Embodiment~\cite{BrohanEtAl2022_OpenX}
& SR
& Robot + Image
& \makecell[l]{Robotics,\\cross-embodiment policy} 
& \makecell[l]{RT-2-X~\cite{BrohanEtAl2022_OpenX}} 
& 2023 \\

RLBench~\cite{RLBench2020}
& SR
& Robot + Image
& \makecell[l]{Robotics,\\manipulation tasks}
& \makecell[l]{/}
& 2020 \\

RoboSuite~\cite{RoboSuite2020}
& SR
& Robot + Image
& \makecell[l]{Robotics,\\simulation environment}
& \makecell[l]{/}
& 2020 \\

ManiSkill2~\cite{ManiSkill2}
& SR
& Robot + Image
& \makecell[l]{Robotics,\\generalizable manipulation}
& \makecell[l]{/}
& 2023 \\

Meta-World~\cite{MetaWorld2019}
& SR
& Robot + State
& \makecell[l]{Robotics,\\multi-task manipulation}
& \makecell[l]{/}
& 2019 \\

CALVIN~\cite{CALVIN2022}
& SR
& Text + Robot + Image
& \makecell[l]{Robotics,\\language-conditioned manipulation}
& \makecell[l]{/}
& 2022 \\

BEHAVIOR~\cite{BEHAVIOR2021}
& SR
& Robot + Image
& \makecell[l]{Robotics,\\household activities}
& \makecell[l]{/}
& 2021 \\

GraspNet-1Billion~\cite{GraspNet2020}
& AP
& RGB-D
& \makecell[l]{Robotics,\\grasp pose prediction}
& \makecell[l]{/}
& 2020 \\

AI2-THOR~\cite{AI2-THOR}
& /
& Image
& \makecell[l]{Navigation,\\embodied simulation} 
& \makecell[l]{/} 
& 2017 \\

RoboTHOR~\cite{RoboTHOR}
& Success Rate \& SPL
& Image
& \makecell[l]{Navigation,\\Sim2Real transfer} 
& \makecell[l]{/} 
& 2020 \\

Gibson~\cite{Gibson}
& Interactive Navigation Score
& Image
& \makecell[l]{Navigation,\\physics interaction} 
& \makecell[l]{/} 
& 2019 \\

HM3D-OVON~\cite{HM3D-OVON}
& SR \& SPL
& Text + Image
& \makecell[l]{Navigation,\\open vocabulary} 
& \makecell[l]{DAgRL+OD~\cite{HM3D-OVON}} 
& 2024 \\

R2R~\cite{mattersim}
& SR
& Text + Image
& \makecell[l]{Navigation,\\visually grounded} 
& \makecell[l]{/} 
& 2018 \\

RxR~\cite{rxr}
& SR \& SPL
& Text + Image
& \makecell[l]{Navigation,\\multilingual grounding} 
& \makecell[l]{/} 
& 2020 \\

VLN-CE~\cite{VLN-CE}
& SR \& SPL
& Text + Image
& \makecell[l]{Navigation,\\continuous embodiment} 
& \makecell[l]{Cross-Modal\\Attention~\cite{VLN-CE}} 
& 2020 \\

REVERIE~\cite{REVERIE}
& SR \& SPL
& Text + Image
& \makecell[l]{Navigation,\\remote grounding} 
& \makecell[l]{INP~\cite{REVERIE}} 
& 2020 \\

Touchdown~\cite{Touchdown}
& TC \& SPD \& SED
& Text + Image
& \makecell[l]{Navigation,\\real-world scene} 
& \makecell[l]{RConcat~\cite{Touchdown}} 
& 2018 \\

RobotSlang~\cite{RobotSlang}
& TD
& Image + Dialog
& \makecell[l]{Navigation,\\dialog-guided interaction} 
& \makecell[l]{/} 
& 2020 \\

CVDN~\cite{CVDN}
& Goal Progress
& Image + Dialog
& \makecell[l]{Navigation,\\dialog-guided localization} 
& \makecell[l]{/} 
& 2020 \\

UNMuTe~\cite{UNMuTe}
& SR
& Image + Dialog
& \makecell[l]{Navigation,\\dialog-based interaction}
& \makecell[l]{/}
& 2024 \\

NavBench~\cite{NavBench}
& SR
& Control Signals
& \makecell[l]{Navigation,\\unified control} 
& \makecell[l]{/} 
& 2025 \\

CommonRoad~\cite{CommonRoad}
& /
& Maps + Dynamics
& \makecell[l]{Autonomous Driving,\\composable motion} 
& \makecell[l]{/} 
& 2017 \\

nuPlan~\cite{nuplan}
& Common \& Scenario-based
& Image + LiDAR + Maps
& \makecell[l]{Autonomous Driving,\\closed-loop planning} 
& \makecell[l]{/} 
& 2021 \\

nuScenes~\cite{nuscenes}
& NDS \& AP
& Image + LiDAR + RADAR
& \makecell[l]{Autonomous Driving,\\multimodal sensing} 
& \makecell[l]{Megvii~\cite{megvii}} 
& 2020 \\

Waymo~\cite{Waymo}
& APH \& AP
& Image + LiDAR
& \makecell[l]{Autonomous Driving,\\scalable perception} 
& \makecell[l]{/} 
& 2020 \\

CARLA~\cite{CARLA}
& SR
& Image + LiDAR + Radar
& \makecell[l]{Autonomous Driving,\\open simulation} 
& \makecell[l]{/} 
& 2017 \\

AirSim~\cite{AirSim}
& /
& Image + Physics Simulation
& \makecell[l]{Autonomous Driving,\\high-fidelity simulation} 
& \makecell[l]{/} 
& 2018 \\

MotionSC~\cite{MotionSC}
& mIoU
& LiDAR + Maps
& \makecell[l]{Autonomous Driving,\\semantic mapping}
& \makecell[l]{/}
& 2022 \\

Bench2Drive~\cite{Bench2Drive}
& SR \& DS \& Efficiency \& Comfortness
& Image + LiDAR + Radar
& \makecell[l]{Autonomous Driving,\\closed-loop assessment} 
& \makecell[l]{DriveAdapter~\cite{driveadapter}} 
& 2024 \\

S2R-Bench~\cite{S2R-Bench}
& Acc.
& Image + LiDAR + Radar
& \makecell[l]{Autonomous Driving,\\sim2real gap} 
& \makecell[l]{/} 
& 2025 \\



\end{xltabular}
\endgroup